\documentclass[sigconf]{acmart}
\AtBeginDocument{%
  \providecommand\BibTeX{{%
    \normalfont B\kern-0.5em{\scshape i\kern-0.25em b}\kern-0.8em\TeX}}}

\copyrightyear{2023} 
\acmYear{2023} 
\setcopyright{rightsretained} 
\acmConference[MM '23]{Proceedings of the 31st ACM International Conference on Multimedia}{October 29-November 3, 2023}{Ottawa, ON, Canada}
\acmBooktitle{Proceedings of the 31st ACM International Conference on Multimedia (MM '23), October 29-November 3, 2023, Ottawa, ON, Canada}
\acmDOI{10.1145/3581783.3611984}
\acmISBN{979-8-4007-0108-5/23/10}

\makeatletter
\patchcmd{\maketitle}{\@copyrightpermission}{
   \begin{minipage}{0.3\columnwidth}
     \href{https://creativecommons.org/licenses/by/4.0/}{\includegraphics[width=0.90\textwidth]{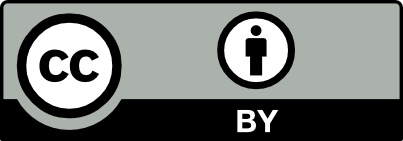}}
   \end{minipage}\hfill
   \begin{minipage}{0.7\columnwidth}
     \href{https://creativecommons.org/licenses/by/4.0/}{This work is licensed under a Creative Commons Attribution International 4.0 License.}
   \end{minipage}
   \vspace{5pt}
}{}{}

\makeatother

\usepackage{algorithm}
\usepackage{algorithmic}
\usepackage{multirow}

\usepackage{float}
\usepackage{subfigure}
\usepackage[procnames]{listings}
\usepackage{stfloats}
\usepackage[toc,page]{appendix}

\begin{document}


\title{Federated Learning with Label-Masking Distillation}

\definecolor{mygray}{gray}{.7}
\def\ie{{\em i.e.}}
\def\eg{{\em e.g.}}
\def\etal{{\em et al.}}

\graphicspath{{./fig/}}

\newcommand{\figref}[1]{Fig. \ref{#1}}
\newcommand{\tabref}[1]{Tab. \ref{#1}}
\newcommand{\equref}[1]{Eq. \ref{#1}}
\newcommand{\secref}[1]{Sec. \ref{#1}}
\newcommand{\algref}[1]{Alg. \ref{#1}}
\newcommand{\myPara}[1]{\vspace{.05in}\noindent\textbf{#1}}
\newcommand{\todo}[1]{\textcolor{red}{\bf [#1]}}
\newcommand{\bl}[1]{\textbf{#1}}
\newcommand{\il}[1]{\textit{#1}}
\newcommand{\mc}[1]{\mathcal{#1}}
\newcommand{\mb}[1]{\mathbf{#1}}
\newcommand{\bul}[1]{\underline{\textbf{#1}}}
\newcommand{\bm}[1]{\mbox{\boldmath{$#1$}}}

\author{Jianghu Lu}
\affiliation{%
  \institution{Institute of Information Engineering, Chinese Academy of Sciences \\School of Cyber Security, UCAS}
}
\email{lujianghu@iie.ac.cn}

\author{Shikun Li}
\affiliation{%
  \institution{Institute of Information Engineering, Chinese Academy of Sciences \\School of Cyber Security, UCAS}
}
\email{lishikun@iie.ac.cn}

\author{Kexin Bao}
\affiliation{%
  \institution{Institute of Information Engineering, Chinese Academy of Sciences \\School of Cyber Security, UCAS}
}
\email{baokexin@iie.ac.cn}

\author{Pengju Wang}
\affiliation{%
  \institution{Institute of Information Engineering, Chinese Academy of Sciences \\School of Cyber Security, UCAS}
}
\email{wangpengju@iie.ac.cn}

\author{Zhenxing Qian}
\affiliation{%
  \institution{School of Computer Science, Fudan University}
}
\email{zxqian@fudan.edu.cn}

\author{Shiming Ge}\authornote{Shiming Ge is the corresponding author (geshiming@iie.ac.cn).}
\affiliation{%
  \institution{Institute of Information Engineering, Chinese Academy of Sciences \\School of Cyber Security, UCAS}
}
\email{geshiming@iie.ac.cn}

\renewcommand{\shortauthors}{Jianghu Lu et al.}


\begin{abstract}
Federated learning provides a privacy-preserving manner to collaboratively train models on data distributed over multiple local clients via the coordination of a global server. In this paper, we focus on label distribution skew in federated learning, where due to the different user behavior of the client, label distributions between different clients are significantly different. When faced with such cases, most existing methods will lead to a suboptimal optimization due to the inadequate utilization of label distribution information in clients. Inspired by this, we propose a label-masking distillation approach termed \emph{FedLMD} to facilitate federated learning via perceiving the various label distributions of each client. We classify the labels into majority and minority labels based on the number of examples per class during training. The client model learns the knowledge of majority labels from local data. The process of distillation masks out the predictions of majority labels from the global model, so that it can focus more on preserving the minority label knowledge of the client. A series of experiments show that the proposed approach can achieve state-of-the-art performance in various cases. Moreover, considering the limited resources of the clients, we propose a variant \emph{FedLMD-Tf} that does not require an additional teacher, which outperforms previous lightweight approaches without increasing computational costs. Our code is available at https://github.com/wnma3mz/FedLMD.
\end{abstract}

\ccsdesc[500]{Computing methodologies~Distributed algorithms}

\keywords{Federated Learning, Knowledge Distillation}


\maketitle

\begin{figure}[ht]
    \centering\includegraphics[width=1.0\linewidth]{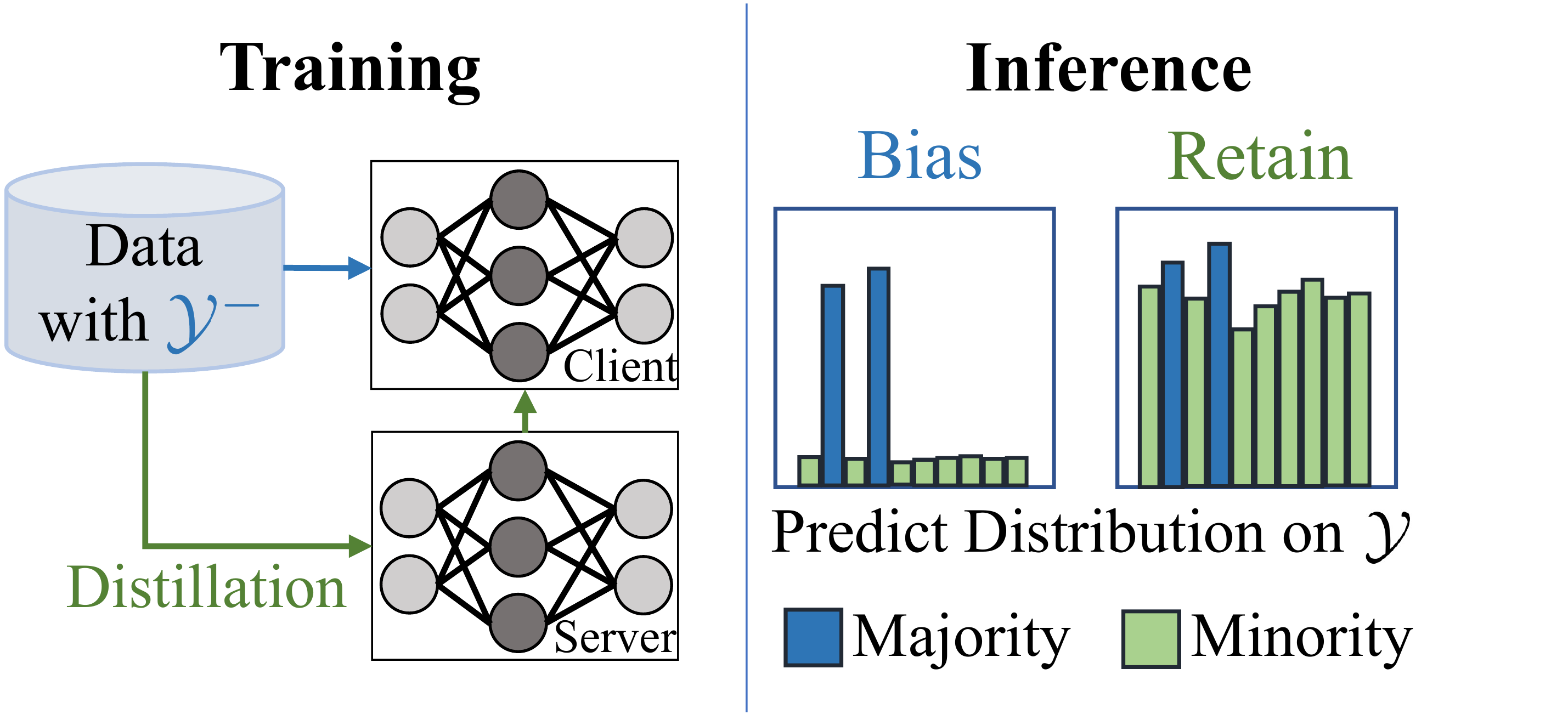}
    \caption{The model trained on the private dataset of a client with partial class labels $\mc{Y}^{-}$ is generally biased to $\mc{Y}^{-}$ due to knowledge missing over complete class labels $\mathcal{Y}$. Our FedLMD method proposes to alleviate it by utilizing the global model from the server to retain the knowledge of \textbf{minority labels} $\mathcal{Y}\backslash\mc{Y}^{-}$.}
    \label{fig:introduce}
\end{figure}

\section{Introduction}

The development of multimedia technology and its various emerging commercial applications have sparked global discussions on the ethics of artificial intelligence~\cite{li2020invisiblefl}. Among these discussions, privacy and security issues have become a key concern for society~\cite{mothukuri2021survey}. Artificial intelligence technology relies heavily on user data uploaded to central servers, which could lead to the leakage of sensitive personal data~\cite{DBLP:journals/csur/NguyenPPDSLDH23}. The centralized collection and use of massive personal data pose serious threats to individual privacy. Once the data are breached or misused, the consequences can be devastating. Additionally, countries worldwide have enacted laws and regulations, such as the European Union's General Data Protection Regulation (GDPR), to restrict such behavior~\cite{philipp2016gdpr,10.2307/j.ctvjghvnn}. Therefore, the multimedia field needs to improve the centralized model training method to gain public recognition and address such concerns.

Federated learning (FL)~\cite{fedavg} has been proposed to provide a feasible solution to jointly train models on distributed data from multiple parties or clients in a privacy-preserving manner. It generally applies a server as coordinator to communicate parameters (gradients or weights of the model) between each client and server, realizing the knowledge sharing rather than data among clients. Since the data only stays local, it is considered to be a privacy-preserving algorithm. It has shown promising results in multimedia applications such as person re-identification~\cite{10.1145/3394171.3413814,DBLP:conf/mm/Zhuang0Z21}, medical images~\cite{liu2021feddg,li2019privacy}, emotion prediction~\cite{10.1145/3503161.3548278} and deepfake detection~\cite{10.1145/3501814}.

In classical FL algorithm FedAvg, the uploaded model parameters are weighted and averaged to implicitly exchange the knowledge of each client. It can work well when the data distributions are identical in clients. However, the realistic data distribution usually is different across clients~\cite{kairouz2019advances}, \ie, non-independent isodistribution (Non-IID). It means that the optimization goals of various client models are much different, and the server-side model is much more difficult to optimize, and may even fail to converge~\cite{fedprox}. In this paper, we focus on a more specific case, \ie, label distribution skew. For instance, diseases can be simply divided into several class labels according to severity, and small clinics in rural areas usually have more examples of minor diseases but fewer or no examples of severe diseases compared to large hospitals. For convenience, we call this realistic scenario the label heterogeneity case.

To address the challenge, some researchers improve FedAvg in terms of weight assignment during aggregation and model aggregation way on the server-side~\cite{wang_federated_2020,lin_ensemble_2020,jeong2018communication,DBLP:journals/corr/abs-1909-06335}. While compared to server-side optimization, client-side optimization is often more effective and straightforward because the data resides on the client-side. The existing client-side methods usually regularize the constraints on the model output or the model parameters themselves~\cite{fedprox,feddyn,fednova,moon,fedntd,xu2022acceleration}. Although these methods can alleviate the challenge in a certain extent, they don't effectively utilize the useful information of varying label distributions in clients under large label heterogeneity, leading to a suboptimal optimization. And this information is crucial and determines the severity of label heterogeneity. Thus, it is necessary to explore an effective solution that can address a key problem: how to exploit the information of label distributions in various clients to perform stable and effective FL?

By revisiting the training process on a particular client in the classical FL~\figref{fig:introduce}, the learned model is prone to be biased toward the majority class labels and forget the absent (or minority) class labels under the label heterogeneity case. In order for the model to learn about minority labels without additional communication, we propose an approach named \textbf{L}abel-\textbf{M}asking \textbf{D}istillation for federated learning (FedLMD) via perceiving the label distributions of each client. The knowledge distillation (KD) has been shown to extract dark knowledge from models and thus reduce the risk of catastrophic forget in FL~\cite{fedntd,fedgkd,he2022class,he2022learning}. As in the previous study, we use the local model as the student, while the global model is updated based on multiple client models. Thus, it is considered as the teacher with more comprehensive label knowledge. To achieve a more effective distillation process, we employ label masking distillation on the client-side model. We classify the labels into majority and minority labels based on the number of examples per label during training. The model can easily learn the knowledge of the majority labels, because of they have sufficient samples. However, the knowledge of the minority labels is prone to being forgotten by the model~\cite{fedntd}. Therefore, to preserve minority knowledge in the model, we only distilled the minority part of the global model to the client-side model. Specifically, we decouple the logits of the global model into two parts: majority and minority, and mask out the majority part of the global logits. Overall, the client-side model learns the knowledge from two sources: majority from the local data and minority from the global model. 

When FL is deployed in real-world applications, the client-side resources have to be seriously considered~\cite{shahid2021communication}. Therefore, we further optimize the computational cost and storage space of the proposed approach. We found that a teacher model with poor performance can still help local models in FL. So we replace the teacher logits with a fixed vector, as demonstrated in~\cite{tfkd}. Since it does not require an additional teacher model, we named it FedLMD-Tf.

In summary, our main contributions are three folds.

\begin{itemize}
    \item We revisit the problem caused by label heterogeneity through a simple experiment and find that the main reason why local models are prone to be biased is the lack of supervision information from minority labels.
    \item We propose FedLMD under the label heterogeneity case. By decoupling the logits of the teacher model and masking out the majority part, the proposed approach is able to retain the forgotten label knowledge for clients by distilling knowledge from the minority part.
    \item We conduct a series of sufficient experiments to show that FedLMD outperforms the state-of-the-art methods on classification accuracy and convergence speed. We also propose FedLMD-Tf which consistently outperforms previous lightweight federated learning methods. 
\end{itemize}

\section{Related Work}
\myPara{Federated Learning on Non-IID Data.}
One of FL's current significant challenges, data heterogeneity, can lead to difficulties in model convergence~\cite{li2020federated}. The optimization can be done from the server-side and the client-side respectively. For server-side optimization, they focus on improving the robustness of the global model by improving the aggregation method~\cite{wang_federated_2020,DBLP:journals/corr/abs-1909-06335,DBLP:conf/icml/YuRMK20,zhang2022fine,DBLP:journals/tpds/WuYW21,DBLP:conf/mm/QiZYZX22}.

The optimization in client focuses on constraining model update to avoid catastrophic forgetting. FedProx~\cite{fedprox} constrains the optimization of local model by computing L2 loss between the local model and global model parameters. Similarly, FedDyn~\cite{feddyn} and FedCurv~\cite{fedcurv} are improved based on the relationship between the model parameters. SCAFFOLD~\cite{scaffold} corrects the local updates by introducing control variates and they are also updated by each client during local training. On this basis, FedNova~\cite{fednova} achieved automatically adjusts the aggregated weight and effective local steps according to the local progress. Unlike these methods, FedRS~\cite{fedrs} adds the scaling factor to SoftMax function using information about the distribution of the data to restrain the update of the parameters of the constrained model updates to the missing classes. 

\myPara{Knowledge Distillation in Federated Learning.}
KD~\cite{hintonDistillingKnowledgeNeural2015} is considered to be able to extract dark knowledge from the teacher. It can be optimized from both server-side and client-side perspectives. Some researchers exploit the feature of multiple models on the server-side of FL to perform integrated multi-teacher KD~\cite{lin_ensemble_2020,sattler2021tnnls,chen_fedbe_2021,DBLP:journals/corr/abs-2104-00352,DBLP:journals/corr/abs-2109-14611,DBLP:journals/corr/abs-2107-00051,sui2020emnlp,gong2021iccv,cho2022heterogeneous}. 

From the perspective of client-side, some studies use data-free KD to expand the local dataset to ensure that the model has access to sufficient data examples during training~\cite{zhu2021icml,zhang2022fine}. However, they cause additional communication overhead and may also result in privacy leakage. Alternatively, KD can improve the performance of the local model by extracting the dark knowledge of the global model. MOON~\cite{moon} constrains the model training by constructing the contrastive loss between the local model and the global model. FedNTD~\cite{fedntd} mitigates the catastrophic forgetting of the global model by removing the target label when the global model is used as a teacher-distilled local student model. While they effectively mitigate the challenge of data heterogeneity and do not introduce additional communication overhead as well as privacy risks, they impose additional computational overhead on the client-side.

In particular, it should be noted that FedNTD~\cite{fedntd} is the most similar to our approach. FedNTD preserves global knowledge, while our approach focuses more on preserving minority label knowledge corresponding to the forgetting of each client. Unlike FedNTD, which only masks out the target class in the teacher model output, we mask out the locally majority labels in the teacher model output from the perspective of label distribution. This achieves more effective knowledge retention. And considering the problem of limited client-side resources, we update the proposed approach to the lightweight version with no additional overhead.

\section{Challenge Revisiting}

\begin{figure}[htbp]
    \centering
    \begin{minipage}{1.0\linewidth}
        \centerline{\includegraphics[width=1.\linewidth]{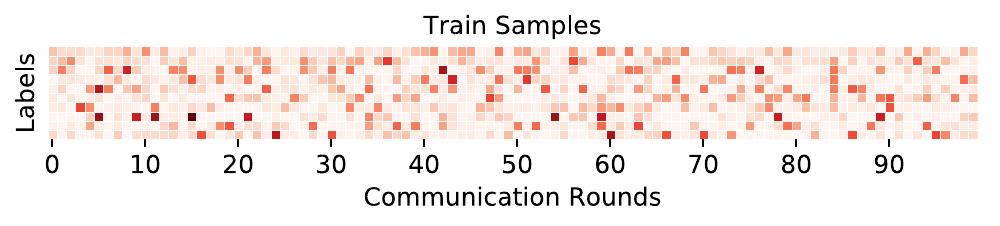}}
        \centerline{\includegraphics[width=1.\linewidth]{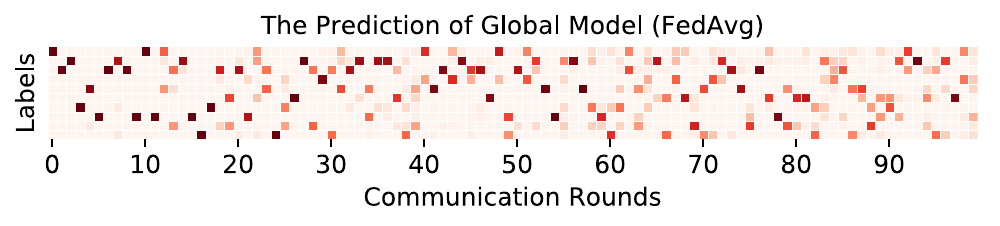}}
        \centerline{\includegraphics[width=1.\linewidth]{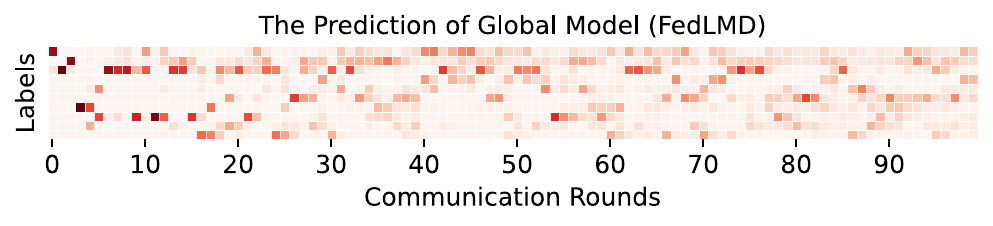}}
    \end{minipage}
    \caption{The label distribution of the training examples (Top), the prediction distribution of the FedAvg (Middle), and the prediction distribution of the FedLMD (Bottom) under different communication rounds.}
    \label{fig:sample_pred}
\end{figure}

To better understand the challenge caused by label heterogeneity, we first experimentally revisit the problem encountered by FedAvg during the training process\footnote{The specific experimental setup can be found at~\secref{para:implement} and we only show the results of the first 100 rounds here.}. The results are shown in~\figref{fig:sample_pred}, where the darker the color is, the greater the number of samples for the corresponding label is.~\figref{fig:sample_pred} (Top), we present the total number of training examples for each label in the uploaded clients under different communication rounds, which clearly shows that the label distribution varies a lot during training. 

~\figref{fig:sample_pred} (Middle), we show the prediction distribution of the FedAvg method under different rounds, which reflects the instability of its optimization. It can be noticed that the class labels with the most training examples severely affect the prediction distribution, making the model biased toward the majority of class labels and forgetting the minority class labels. Specifically, in the 9-th round, when class label 7 has the most examples, then the prediction distribution of the model is largely biased toward class label 7. Although the model is relatively less affected by the heterogeneity at the later stage of training (\eg, after 100 rounds), the bias toward majority labels still exists. Therefore, it can be found that the main reason why local models are prone to be biased under such cases is the \emph{lack of supervision information from minority class labels}, which inspires us to introduce the information of minority class labels into supervision. 

By perceiving the label distributions, as shown in~\figref{fig:sample_pred} (Bottom), our FedLMD approach can well resist the bias of majority labels, leading to stable and effective optimization. It can see that the color depth of different labels tends to be the same at the later stage of training. It means that the prediction label distribution achieved by our FedLMD approach is close to the uniform distribution.

\section{Proposed Method}

\begin{figure*}[!ht]
    \centering
    \includegraphics[width=1.0\linewidth]{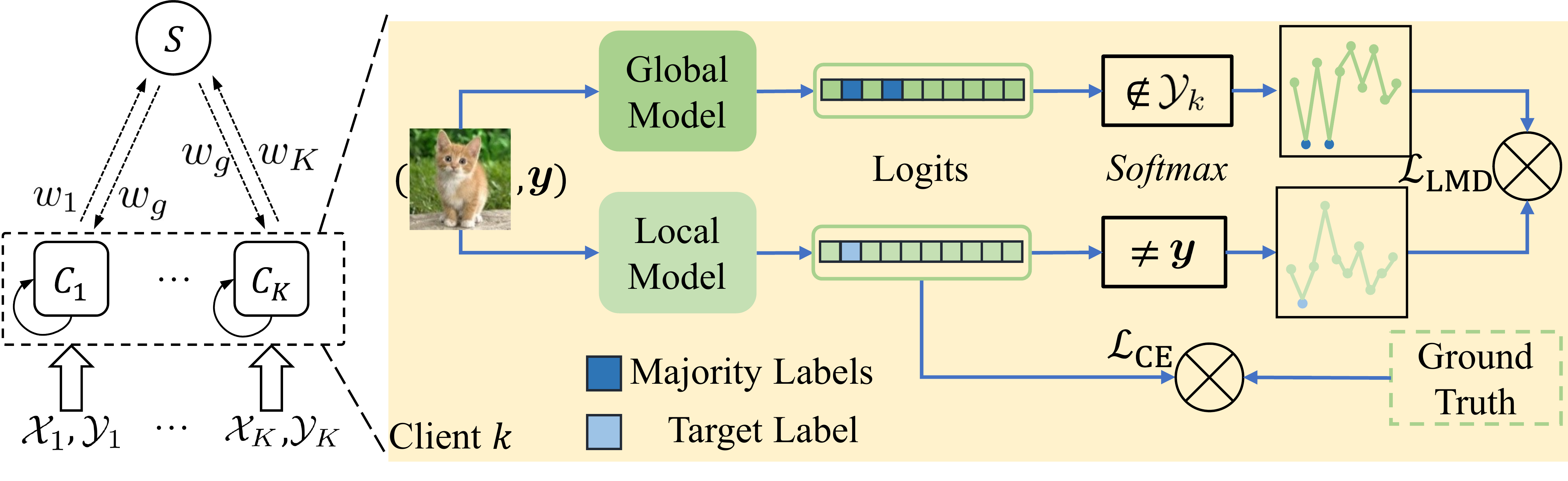}
    \caption{The framework of our approach. For the aggregation process, for the uploaded weight $w_1, ... ,w_K$ of the model are calculated as weighted averages to obtain $w_g$. For each client, the training loss is the combination of the cross-entropy loss $\mathcal{L}_{\rm CE}$ for learning from local data and the label-masking distillation loss $\mathcal{L}_{\rm LMD}$ for distilling from the global model.}
    \label{fig:method}
\end{figure*}

\subsection{Problem Setting}

We consider a classical supervised FL system that contains a server and $K$ clients. For the $k$-th client, it has a local dataset $\mathcal{D}_{k}=\{\left(\bm{x}_i,\bm{y}_i\right)\}_{i=1}^{n_k}$, where as $\left(\bm{x}_i, \bm{y}_i\right) \in \left(\mathcal{X}_{k}, \mathcal{Y}_{k}\right)$ and the weight parameters of model is $w_k$. The goal of FL is to obtain a global model by jointly training all clients as follows:
\begin{equation}
\min\limits_{w_g}\frac{1}{|\mathcal{K}|}\sum_{k\in\mathcal{K}}\mathcal{L}_k (w_g; \mathcal{D}_k)
\end{equation}

where $w_g$ is the weight of the global model and $\mathcal{L}_k$ is the loss function for training the $k$th client model. On the server-side, the FL system aggregates all uploaded model weights. In each communication round, the clients are specified in $\mathcal{K}$ to train and upload parameters, where $|\mathcal{K}|$ is the number of models to upload. 

As mentioned before, label heterogeneity in clients can make the local model biased to the majority labels, leading to unstable and poor optimization. Our goal is to facilitate stable and effective FL via perceiving the various label distributions of each client.

\subsection{Label-Masking Distillation}
First of all, we divide all class labels into majority labels and minority labels. When $n_{k,y} >= \frac{n_{k}}{n_{k,y}}$, class label $y$ is a majority label in the $k$-th client. The $n_k$ is the total number of samples for all classes and $n_{k,y}$ is the number of samples for class $y$ of the $k$-th client. In this section, we assume the majority labels are all in $\mathcal{Y}_k$ on $k$-th client for the sake of convenient expression. 

For a training example $(\bm{x},\bm{y})\in \left(\mathcal{X}_{k}, \mathcal{Y}_{k}\right)$, let the output of the $k$-th local model as $p_k$, the output of the global model as $p_g$, and $\mathbf{1}_{{y}}$ is the one-hot vector form of $\bm{y}$. KD~\cite{hintonDistillingKnowledgeNeural2015} is to achieve dark knowledge transfer by making the output of the student mimic the output of the teacher. Since the global model is updated based on multiple client models, we use the local model as the student and regard the global model as the teacher with more comprehensive label knowledge, the $k$-th client's loss can achieve the aim of knowledge retention as follows:
\begin{equation}
    \label{equation:KD}
    \begin{aligned}
        \mathcal{L}_k & = \mathcal{L}_{\rm CE}(p_k, \mathbf{1}_{{y}}) + \beta \mathcal{L}_{\rm KD} (p_k, p_g),
    \end{aligned}
\end{equation}
where $\mathcal{L}_{\rm CE}$ is the cross-entropy loss for learning the majority labels knowledge, and $\mathcal{L}_{\rm KD}$ is the distillation loss for retaining all the labels knowledge. Here, we fix the weight of $\mathcal{L}_{\rm CE}$ to 1 and $\beta$ is used as a weighting factor to control the distillation loss.

Although the $\mathcal{L}_{\rm KD}$ can learn from $k$-th data and assist the bias toward minority labels, it performs the regularization without considering the varying label distributions across clients, leading to a suboptimal optimization. Hence, we improve it by enhancing the KD for minority labels via perceiving the label distributions. We decouple the logits of the global model into two parts: majority and minority. The majority part of teacher logits corresponds to majority labels, and naturally, the minority part corresponds to minority labels. We focus on the minority part of the teacher logits for distillation by masking out the majority part. Because of the majority labels knowledge can be learned from $\mathcal{L}_{\rm CE}$. This leads to a modified teacher distribution $p^{\prime}_g$ as: 
\begin{equation}
    p^{\prime}_g(i|\bm{x}) =\left\{
    \begin{array}{lcl}
        \frac{\exp(\bm{z}_{g,i}/\tau)}{\sum_{i=1,i\notin \mathcal{Y}_k}^{C}\exp(\bm{z}_{g,i}/\tau)} & , & {i\notin \mathcal{Y}_k} \\
        0                                                                       & , & {i\in \mathcal{Y}_k}
    \end{array} \right.,
    \label{equ:pg}
\end{equation}
where $\bm{z}_{g,i}$ is the logits of the global model for $i$-th class label, and $\tau$ is a temperature factor. We mask out the majority labels for teacher logits (set to 0), which encourages the student model to learn from the knowledge of the minority labels or not all labels, and helps prevent forgetting this knowledge. 

For the student model's predictions $p_k$, a straightforward way is to leave it unchanged. However, this leads to a conflict between $\mathcal{L}_{\rm CE}$ and distillation loss. Because of the teacher's logits for the target label is 0 in distillation (\equref{equ:pg}) and the one-hot vector $\mathbf{1}_y$ is 1 in $\mathcal{L}_{\rm CE}$. Therefore, we mask out the target label in the student model to avoid such conflicts. Additionally, for the majority not-target labels, the student's performance can be further improved by learning from negative supervision~\cite{KimYYK19,00030YYXTS20}. Therefore, we modify the distribution from the student model as: 
\begin{equation}
    \label{equation:pl_prime}
    p^{\prime}_k(i|\bm{x}) =\left\{
    \begin{array}{lcl}
        \frac{\exp(\bm{z}_{k,i}/\tau)}{\sum_{i=1,i\neq {y}}^{C}\exp(\bm{z}_{k,i}/\tau)} & , & {i\neq \bm{y}} \\
        0                                                                & , & {i= \bm{y}}
    \end{array} \right.,
\end{equation}
where $\bm{z}_{k,i}$ is the logits of the $k$-th client model for $i$-th class label.

Then, the improved loss can be proposed as follows:
\begin{equation}
	\label{equation:LMD}
	\begin{aligned}
		\mathcal{L}_k & = \mathcal{L}_{\rm CE}(p_k, \mathbf{1}_{{y}}) + \beta \mathcal{L}_{\rm LMD} (p^{\prime}_k, p^{\prime}_g),
	\end{aligned}
\end{equation}
where the label-masking distillation loss $\mathcal{L}_{\rm LMD}$ is defined as the Kullback-Leibler divergence between $p^{\prime}_k$ and $p^{\prime}_g$:
\begin{equation}
	\begin{aligned}
		\mathcal{L}_{\rm LMD}(p^{\prime}_k, p^{\prime}_g)=-\sum_{i=1}^{C}p^{\prime}_g(i|\bm{x})\log\frac{p^{\prime}_k(i|\bm{x})}{p^{\prime}_g(i|\bm{x})}.
	\end{aligned}
\end{equation}

And the framework of FedLMD can be seen in \figref{fig:method}.

\subsection{Teacher-free Variant}
In practical scenarios, the clients may have limited storage space and computation resource. FedLMD introduces an extra model for each client, which will undoubtedly increase the hardware overhead of the client. Therefore, we consider dropping the teacher model to avoid the cost.

\label{sec:lmd2tf}
The teacher model in distillation generally needs to be pretrained so that they can better provide knowledge to the student model. However, FL is an online learning,~\ie, the teacher model does not have good performance in the beginning stage. Inspired by~\cite{tfkd}, we treat distillation as the label smoothing (LS) regularization by introducing a fixed minority label distribution to replace the output of the teacher model. Specifically, we replace $p^{\prime}_g$ with $\mu_k$ in~\equref{equation:LMD_tf} as follows:
\begin{equation}
\label{equation:LMD_tf}
\begin{aligned}
    \mathcal{L}_k & = \mathcal{L}_{\rm CE}(p_k, \mathbf{1}_{{y}}) + \beta \mathcal{L}_{\rm LMD} (p^{\prime}_k, \mu_k).
\end{aligned}
\end{equation}
The fixed minority label distribution for $k$-th client is
\begin{equation}
    \label{equation:mu}
    \mu_k(i) =\left\{
    \begin{array}{lcl}
        1/(C-C_k) & , & {i\notin \mathcal{Y}_k} \\
        0         & , & {i\in \mathcal{Y}_k}.
    \end{array} \right.,
\end{equation}
where $C$ and $C_k$ denote the total number of class labels, the number of majority labels for the $k-$th client, respectively. 

As this method does not require the teacher model, it is named FedLMD-Tf (Teacher-free). In this way, such a lightweight version does not increase computation, communication overhead and privacy risk, and can achieve much better performance via perceiving the label distributions. The detailed training is shown in~\algref{algorithm:FedLMD}.

\begin{algorithm}[tb]
    \caption{FedLMD and FedLMD-Tf. $T$ is the number of communication rounds, $E$ the local epochs, and $\eta$ the learning rate. Indices $k$ denote $K$ clients with local dataset $\mathcal{D}_{k}$; $w_g^t$ and $w_k^t$ are the global and $k$-th client model weights at round $t$; $\mathcal{K}$ is the set of selected clients per round.}
    \label{algorithm:FedLMD}
    \begin{algorithmic}[1]
        \STATE Initialization: $w^0_g$\
        \FOR{each round $t=1,2,...,T$}
            \STATE Broadcasts $w^{t}_k \leftarrow w^{t-1}_g (k\in[1,...,K])$
        \STATE $\mathcal{K}\leftarrow$ a random subset of the $K$ clients.\
        \FOR{each client $k\in \mathcal{K}$ \textbf{in parallel}}
        \FOR{local training steps $e = 1,...,E$}
        \STATE // Using~\equref{equation:LMD} for FedLMD or~\equref{equation:LMD_tf} for FedLMD-Tf \
        \STATE $w^{t}_{k}= w^{t}_k - \eta \nabla_{w}\mathcal{L}_{k}(w^{t}_k, \mathcal{D}_{k}, w^{t-1}_g)$ 
        \ENDFOR
        \ENDFOR
        \STATE Upload $w^{t}_{k}$ ($k \in \mathcal{K}$) to the server
    
        \STATE $w^{t}_g=\frac{1}{|\mathcal{K}|}\sum\limits_{k\in \mathcal{K}}w_{k}^{t}$\
        \ENDFOR
    \end{algorithmic}
\end{algorithm}

\section{Experiments}

\subsection{Experimental Setup}

\myPara{Baselines.}~The methods we compare focus on the traditional FedAvg~\cite{fedavg} and on algorithms that are client-side improved for data heterogeneity problems (FedProx~\cite{fedprox}, FedCurv~\cite{fedcurv}, SCAFFOLD~\cite{scaffold}, FedNova~\cite{fednova},  FedRS~\cite{fedrs}, MOON~\cite{moon}, FedNTD~\cite{fedntd}). For some experiments, we have selected only a few important methods (FedAvg, FedCurv, FedProx, FedNTD) for comparison. All methods are replicated based on the PyTorch framework(1.10.0+cu113) and experimented on RTX 3090 and Intel(R) Xeon(R) Silver 4214R CPU @ 2.40GHz. 


\myPara{Datasets.}~For a fair comparison, we decided to use the same experimental setup as~\cite{fedntd}. We used four datasets, MNIST~\cite{726791}, CINIC-10~\cite{darlow2018cinic}, CIFAR-10 and CIFAR100~\cite{krizhevsky2009learning}. The dataset is sliced using two Non-IID partition strategies respectively: 1) \textbf{Sharding}~\cite{fedavg}: The data is sliced according to labels, and the sliced data is called a slice. Each slice has the same number of examples, and the degree of data heterogeneity is determined by the number of slices $s$ each client has. We set s to MNIST ($s = 2$), CIFAR-10 ($s =2,3,5,10$), CIFAR-100 ($s = 10$), and CINIC-10 ($s = 2$). 2) \textbf{Latent Dirichlet Allocation (LDA)}~\cite{moon}: The dataset is sliced by dirichlet sampling, which provides unbalanced labels and unbalanced examples for each client. And the degree of data heterogeneity of different clients is determined by controlling $\alpha$. We set $\alpha$ as MNIST ($\alpha=0.1$), CIFAR-10 ($\alpha=0.05, 0.1, 0.3, 0.5$), CIFAR-100 ($\alpha=0.1$), and CINIC-10 ($\alpha=0.1$). 


\label{para:implement}
\myPara{Implementation.}~For a fair comparison, we use a network model with two convolution layers followed by max-pooling layers, and two fully-connected layers for all methods. The cross-entropy loss and the SGD optimizer are adopted. The learning rate is set to 0.01 and it decays with a factor of 0.99 at each communication round. The weight decay is set to 1e-5 and the SGD momentum is set to 0.9. The batch size is set to 50. For data augmentation, we employ techniques such as random cropping, random horizontal flipping, and normalization. Note that our default experimental dataset is CIFAR-10 ($\alpha=0.05$) unless specified. 

For the FL task, we set some additional hyperparameters. Referring to the settings of previous studies, we set the number of clients $K=100$, the number of local training epochs $E=5$, the communication rounds $T=200$, and randomly select $|\mathcal{K}|=10$ clients per round. 

In all of the experiments, we conduct a grid search on the parameters of each method to determine the optimal performance. After each communication round, we evaluate the global model on the test dataset and select the best test accuracy as the result display. 



\subsection{Improvement with Knowledge Distillation}
\label{sec:kd}

In this subsection, we utilize and improve KD to alleviate the situation that the model is prone to be biased toward majority labels under the label heterogeneity case.

\begin{figure}[H]
    \centering
    \includegraphics[width=\linewidth]{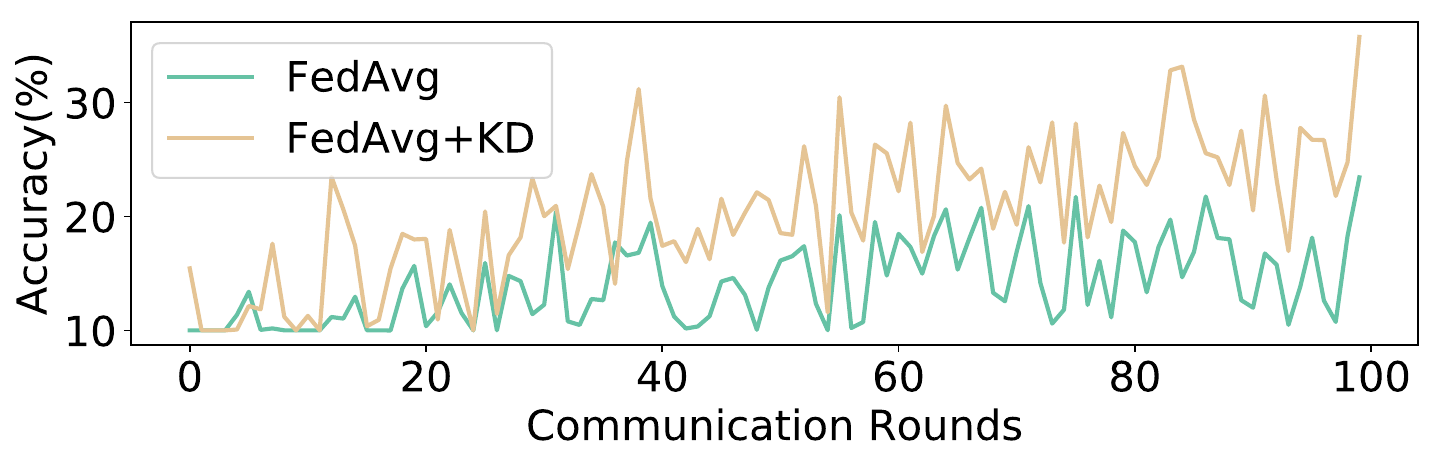}
    \caption{The effect of knowledge distillation in FedAvg on CIFAR-10 ($\alpha=0.05$).}
    \label{fig:exp_LMD}
\end{figure}

First of all, we briefly compared the change in FedAvg accuracy after applying distillation and the results are shown in~\figref{fig:exp_LMD}. We found that KD can help FedAvg alleviate the label heterogeneity problem. However, the traditional KD treats all labels in the same way, which affects the effectiveness of dark knowledge transfer. Therefore, FedNTD only selects non-target labels for distillation. And our proposed FedLMD goes one step further by masking out the majority labels in the output of the teacher model, \ie, selecting minority labels for distillation. From~\tabref{tab:exp_kd}, we can find that FedLMD has significant superiority.

Moreover, the teacher is often assumed to be a well-pretrained model in KD. But the global model performs poorly at the beginning stage in FL. As shown in~\figref{fig:exp_LMD}, the poor global model as a teacher still improved student performance at the beginning of FL. This observation inspired us to discard the teacher model and use a fixed distribution vector as an unreliable teacher to replace it.

\begin{table}[htbp]
    \centering
    \caption{The top-1 test accuracy (\%) on CIFAR-10 under different distillation methods ($\alpha=0.05$).}
    \label{tab:exp_kd}
    \begin{tabular}{l|c|c|c|c}
        \toprule
        Method                        & FedAvg  & FedAvg+KD    & FedNTD  & FedLMD \\ \midrule
        Accuracy                      & 33.02   & 40.46  & 47.01   & \textbf{50.45}  \\\bottomrule
    \end{tabular}
\end{table}

Further, to better understand the effectiveness of our teacher-free distillation, we use LS ($\mu=0.1$) on FedAvg which is similar to the teacher-free distillation. In addition, to be fair, we modified FedNTD to a teacher-free version as well, called FedNTD-Tf, for comparison. From the~\tabref{tab:exp_lsr}, we find that LS and FedNTD-Tf can alleviate the bias of FL. And when we use teacher-free distillation for minority labels, the FedLMD-Tf is further enhanced by being more focused on preserving the minority label knowledge.

\begin{table}[htbp]
    \centering
    \caption{The top-1 test accuracy (\%) on CIFAR-10 under teacher-free (Tf) distillation ($\alpha=0.05$). LS: label smoothing.}
    \label{tab:exp_lsr}
    \begin{tabular}{l|c|c|c|c}
        \toprule
        Method                 & FedAvg & FedAvg+LS   & FedNTD-Tf & FedLMD-Tf   \\ \midrule
        Accuracy               & 33.02   & 42.74   & 41.30 & \textbf{45.08} 	\\ \bottomrule
    \end{tabular}
\end{table}

\subsection{Results on Label Heterogeneity}
In this subsection, we compare FedLMD and FedLMD-Tf with the previous FL methods comprehensively. 

\myPara{Accuracy and Convergence Speed.}~We show the results of our experiments with two different strategies of data partition in~\tabref{tab:sharding_lda_acc}. The effectiveness of the proposed approach is well illustrated by different datasets and the degree of label heterogeneity. Especially in the case of CIFAR-10 ($\alpha=0.05$), it is up to 17\% improvement over FedAvg. FedLMD outperforms the previous results in the vast most of cases, and the performance improvement becomes more and more obvious as the degree of label heterogeneity increases ($\alpha$ or $s$ keeps decreasing). Even though the results in a few cases are not the best, they are still very close to the SOTA baselines. Moreover, we measure the communication rounds required for different methods to reach the top-1 test accuracy of FedAvg, which is used as the evaluation metric for convergence speed~\cite{moon}. As shown in in~\tabref{tab:sharding_lda_acc}, FedLMD clearly converges faster than the other methods. Specifically, in the experiment on MNIST dataset, it achieves 2.47 times speedup against FedAvg. 

\begin{figure*}
    \centering
    \subfigure[CIFAR-10 ($\alpha=0.05$)]{
        \begin{minipage}[t]{0.3\linewidth}
            \centering
            \includegraphics[width=1.0\linewidth]{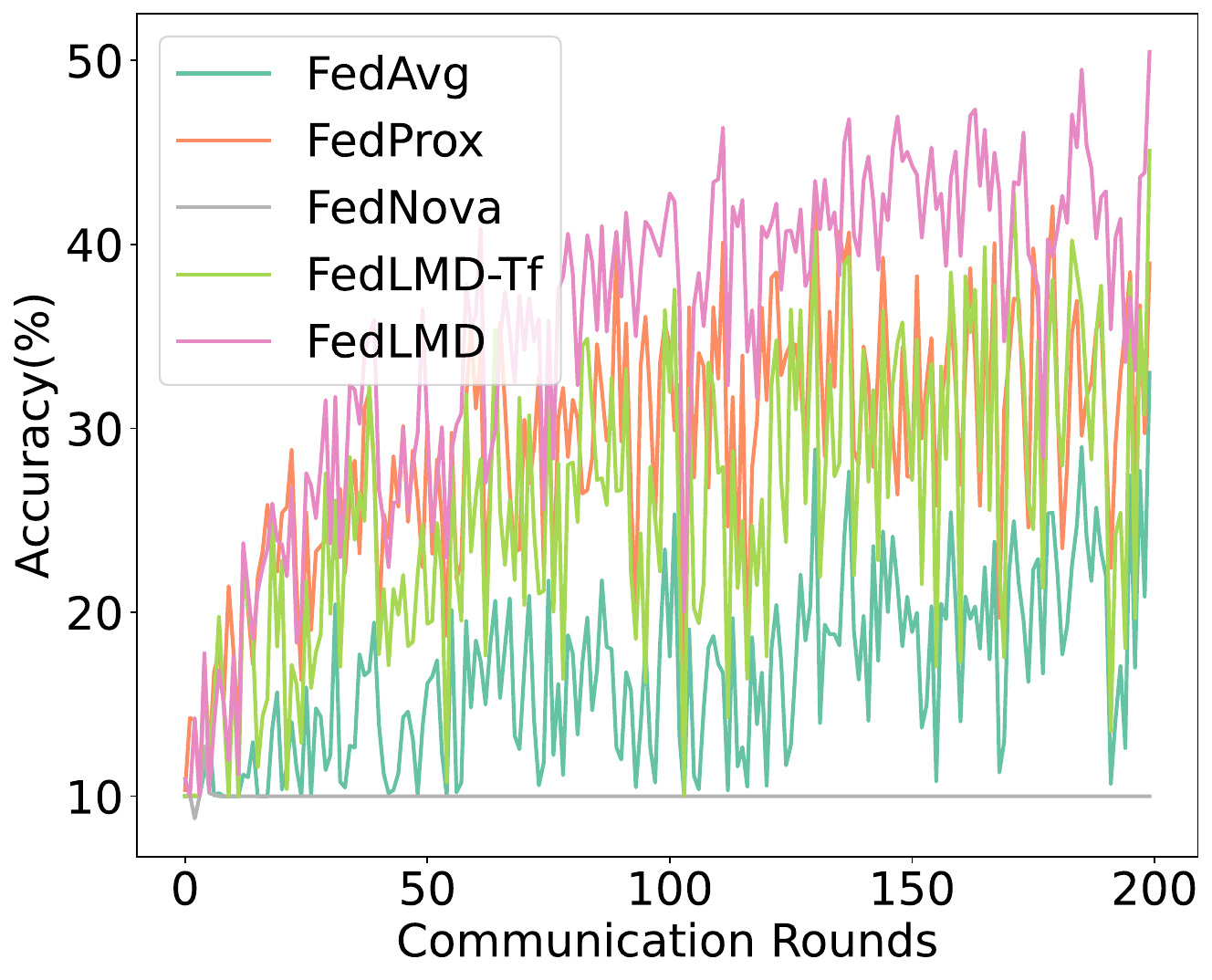}
        \end{minipage}
    }
    \subfigure[CIFAR-10 ($\alpha=0.3$)]{
        \begin{minipage}[t]{0.3\linewidth}
            \centering
            \includegraphics[width=1.0\linewidth]{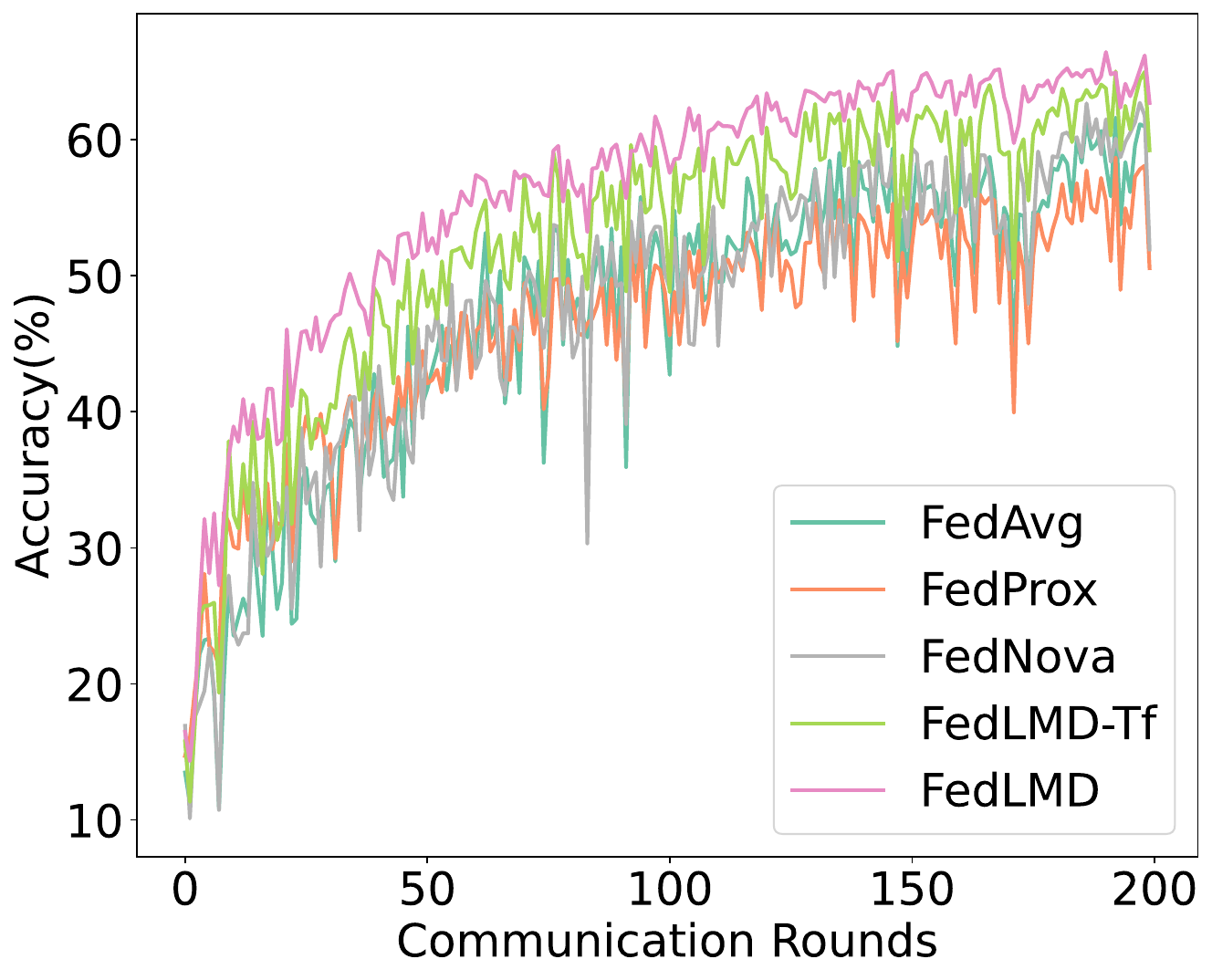}
        \end{minipage}
    }
    \subfigure[CIFAR-10 ($\alpha=0.5$)]{
        \begin{minipage}[t]{0.3\linewidth}
            \centering
            \includegraphics[width=1.0\linewidth]{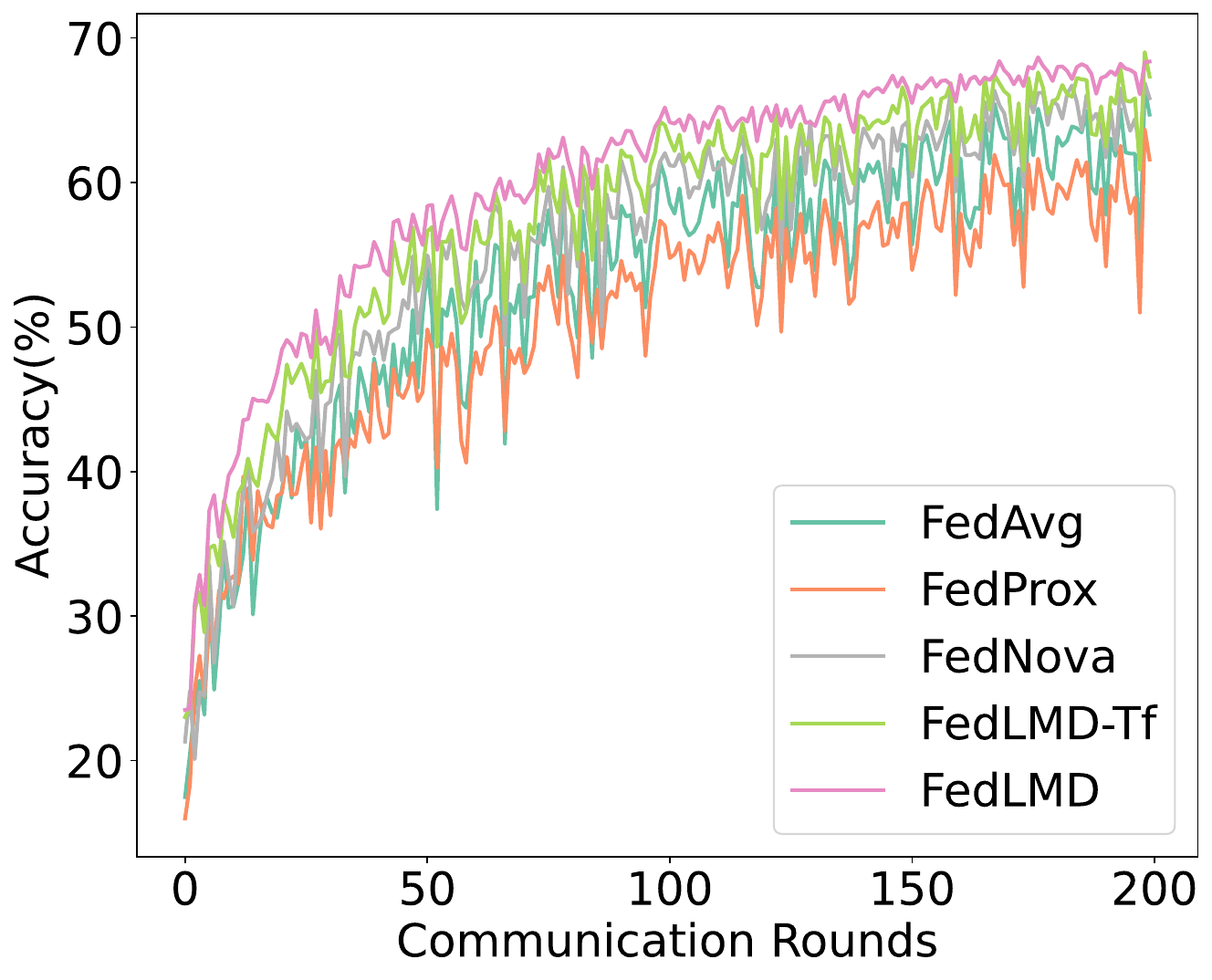}
        \end{minipage}
    }
    \centering
    \caption{The top-1 test accuracy (\%) of different approaches on CIFAR-10 dataset under different communication rounds.}
    \label{fig:comm_effi}
\end{figure*}

We compared the training processes of different methods on the CIFAR-10 dataset. We evaluated their test accuracy on CIFAR-10 under three scenarios: $\alpha=0.05$, $0.3$, and $0.5$. As illustrated in~\figref{fig:comm_effi}, FedLMD exhibited greater stability during training compared to the other methods in each case scenario.

\begin{figure}[htbp]
    \centering
    \includegraphics[width=\linewidth]{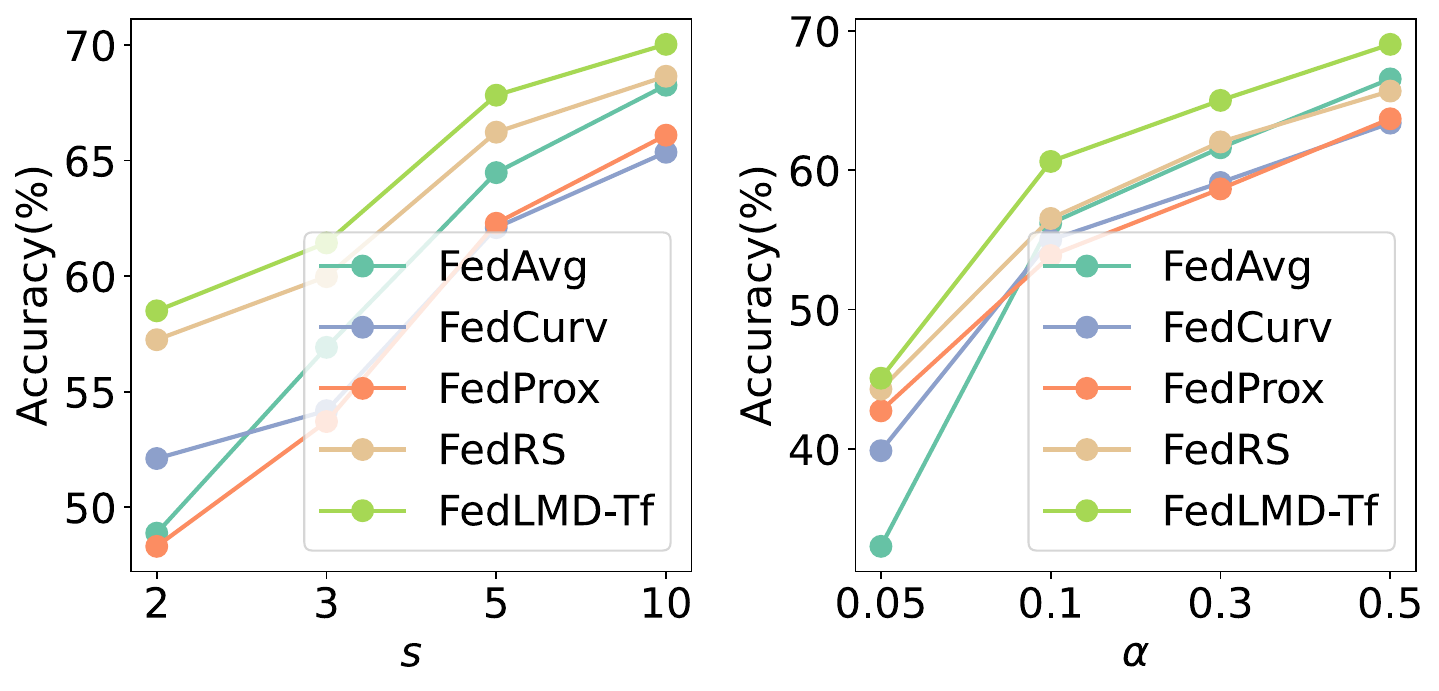}
    \caption{Comparison of the accuracy (\%) of the method without additional computational cost on two partition strategies Sharding (Left) and LDA (Right) of CIFAR-10.}
    \label{fig:exp_abl_main_tf}
\end{figure}

\myPara{Comparison with Light Baselines. }~When FL is deployed on low-power devices, it have to consider the client-side computational costs. Therefore, in such a situation, lightweight FL methods are valuable. Here, we compare the performance of FedLMD-Tf with some previous lightweight approaches on the CIFAR-10 dataset to show its advantage. As shown in~\figref{fig:exp_abl_main_tf}, FedLMD-Tf consistently outperforms other methods under various cases without increasing computational costs. We should additionally note that the size of the vector predefined by FedLMD-Tf on each client depends on the number of class labels $C$ in the FL system.

\begin{table*}
	\centering
	\caption{The top-1 test accuracy (\%) on MNIST, CIFAR-10, CIFAR-100, and CINIC-10. The values in the parentheses are the speedup of the approach computed against FedAvg. If \textit{Failed} is displayed in the parentheses, the method cannot be converged.}
	\label{tab:sharding_lda_acc}
	\resizebox{1.0\textwidth}{!}{
		\begin{tabular}{l|ccccccc}
			\toprule
			\multicolumn{8}{c}{\textbf{\large The Non-IID Partition Strategy: Sharding}}                                                                                                                                                                                                                                                                                                       \\ \midrule

			\multicolumn{1}{l|}{\multirow{2}{*}{\textbf{Method}}} & \multicolumn{1}{c}{\multirow{2}{*}{\textbf{MNIST}}} & \multicolumn{4}{c}{\textbf{CIFAR-10}}  & \multicolumn{1}{l}{\multirow{2}{*}{\textbf{CIFAR-100}}} & \multirow{2}{*}{\textbf{CINIC-10}}                                                                                                                                \\
			\multicolumn{1}{l|}{}                                 & \multicolumn{1}{c}{}                                & $s=2$                                  & $s=3$                                                   & $s=5$                                  & \multicolumn{1}{c}{$s=10$}             &                                        &                                        \\ \midrule
			FedAvg                                                & 85.41 (\textit{1.00}$\times$)                       & 48.88 (\textit{1.00}$\times$)          & 56.92 (\textit{1.00}$\times$)                           & 64.48 (\textit{1.00}$\times$)          & 68.26 (\textit{1.00}$\times$)          & 26.97 (\textit{1.00}$\times$)          & 50.66 (\textit{1.00}$\times$)          \\ \midrule
			FedCurv                                               & 85.08 (\textit{1.00}$\times$)                       & 52.11 (\textit{1.18}$\times$)          & 54.18 (\textit{1.00}$\times$)                           & 62.10 (\textit{1.00}$\times$)          & 65.36 (\textit{1.00}$\times$)          & 24.56 (\textit{1.00}$\times$)          & 49.52 (\textit{1.00}$\times$)          \\
			FedProx                                               & 83.11 (\textit{1.00}$\times$)                       & 48.31 (\textit{1.00}$\times$)          & 53.71 (\textit{1.00}$\times$)                           & 62.29 (\textit{1.00}$\times$)          & 66.10 (\textit{1.00}$\times$)          & 26.88 (\textit{1.00}$\times$)          & 48.51 (\textit{1.00}$\times$)          \\
			FedNova                                               & 85.34 (\textit{1.00}$\times$)                       & 50.69 (\textit{1.07}$\times$)          & 57.98 (\textit{1.03}$\times$)                           & 65.28 (\textit{1.17}$\times$)          & 68.64 (\textit{1.09}$\times$)          & 29.11 (\textit{1.49}$\times$)          & 49.12 (\textit{1.00}$\times$)          \\
			SCAFFOLD                                              & 86.13 (\textit{1.41}$\times$)                       & 54.62 (\textit{1.44}$\times$)          & 40.73 (\textit{1.00}$\times$)                           & 67.25 (\textit{1.59}$\times$)          & \textbf{70.79} (\textit{1.46}$\times$) & 31.92 (\textit{1.79}$\times$)          & 52.89 (\textit{1.54}$\times$)          \\
			MOON                                                  & 85.25 (\textit{1.00}$\times$)                       & 48.40 (\textit{1.00}$\times$)          & 57.01 (\textit{1.27}$\times$)                           & 64.34 (\textit{1.00}$\times$)          & 68.36 (\textit{1.07}$\times$)          & 27.20 (\textit{1.02}$\times$)          & 49.86 (\textit{1.00}$\times$)          \\
			FedRS                                                 & 85.54 (\textit{1.15}$\times$)                       & 57.25 (\textit{2.60}$\times$)          & 59.98 (\textit{1.69}$\times$)                           & 66.23 (\textit{1.41}$\times$)          & 68.65 (\textit{1.07}$\times$)          & 30.24 (\textit{1.82}$\times$)          & 51.99 (\textit{1.17}$\times$)          \\
			FedNTD                                                & 87.76 (\textit{1.98}$\times$)                       & 60.03 (\textit{3.28}$\times$)          & 61.65 (\textit{1.79}$\times$)                           & 68.08 (\textit{2.02}$\times$)          & 70.06 (\textit{1.46}$\times$)          & 32.27 (\textit{2.13}$\times$)          & \textbf{54.14} (\textit{1.69}$\times$) \\ \midrule
			FedLMD                                          & \textbf{88.48} (\textit{2.02}$\times$)              & \textbf{60.76} (\textit{3.64}$\times$) & \textbf{62.44} (\textit{2.04}$\times$)                  & \textbf{69.20} (\textit{2.02}$\times$) & 70.32 (\textit{1.52}$\times$)          & \textbf{32.34} (\textit{2.30}$\times$) & 54.13 (\textit{2.02}$\times$)          \\
			\midrule\midrule
			\multicolumn{8}{c}{\textbf{\large The Non-IID Partition Strategy: LDA}}                                                                                                                                                                                                                                                                                                            \\ \midrule

			\multicolumn{1}{l|}{\multirow{2}{*}{\textbf{Method}}} & \multicolumn{1}{c}{\multirow{2}{*}{\textbf{MNIST}}} & \multicolumn{4}{c}{\textbf{CIFAR-10}}  & \multirow{2}{*}{\textbf{CIFAR-100}}                     & \multirow{2}{*}{\textbf{CINIC-10}}                                                                                                                                \\
			\multicolumn{1}{l|}{}                                 & \multicolumn{1}{c}{}                                & $\alpha=0.05$                          & $\alpha=0.1$                                            & $\alpha=0.3$                           & \multicolumn{1}{c}{$\alpha=0.5$}       &                                        &                                        \\ \midrule
			FedAvg                                                & 85.19 (\textit{1.00}$\times$)                       & 33.02 (\textit{1.00}$\times$)          & 56.19 (\textit{1.00}$\times$)                           & 61.61 (\textit{1.00}$\times$)          & 66.55 (\textit{1.00}$\times$)          & 31.36 (\textit{1.00}$\times$)          & 55.64 (\textit{1.00}$\times$)          \\ \midrule
			FedCurv                                               & 84.76 (\textit{1.00}$\times$)                       & 39.88 (\textit{2.35}$\times$)          & 55.00 (\textit{1.00}$\times$)                           & 59.12 (\textit{1.00}$\times$)          & 63.38 (\textit{1.00}$\times$)          & 29.65 (\textit{1.00}$\times$)          & 54.52 (\textit{1.00}$\times$)          \\
			FedProx                                               & 82.43 (\textit{1.00}$\times$)                       & 42.74 (\textit{3.33}$\times$)          & 53.88 (\textit{1.00}$\times$)                           & 58.66 (\textit{1.00}$\times$)          & 63.69 (\textit{1.00}$\times$)          & 28.44 (\textit{1.00}$\times$)          & 54.06 (\textit{1.00}$\times$)          \\
			FedNova                                               & 77.07 (\textit{1.00}$\times$)                       & 12.71 (\textit{Failed})                & 41.86 (\textit{1.00}$\times$)                           & 62.70 (\textit{1.07}$\times$)          & 66.89 (\textit{1.09}$\times$)          & 32.78 (\textit{1.33}$\times$)          & 34.67 (\textit{1.00}$\times$)          \\
			SCAFFOLD                                              & 81.75 (\textit{1.00}$\times$)                       & 12.34 (\textit{Failed})                & 28.18 (\textit{1.00}$\times$)                           & 63.74 (\textit{1.24}$\times$)          & 68.07 (\textit{1.20}$\times$)          & \textbf{34.69} (\textit{1.53}$\times$) & 25.19 (\textit{1.00}$\times$)          \\
			MOON                                                  & 85.63 (\textit{1.24}$\times$)                       & 34.54 (\textit{1.00}$\times$)          & 56.70 (\textit{1.03}$\times$)                           & 62.21 (\textit{1.07}$\times$)          & 66.43 (\textit{1.00}$\times$)          & 31.49 (\textit{1.01}$\times$)          & 55.80 (\textit{1.04}$\times$)          \\
			FedRS                                                 & 85.37 (\textit{1.22}$\times$)                       & 44.30 (\textit{5.00}$\times$)          & 56.54 (\textit{1.03}$\times$)                           & 62.03 (\textit{1.11}$\times$)          & 65.68 (\textit{1.00}$\times$)          & 31.51 (\textit{1.02}$\times$)          & \textbf{58.21} (\textit{1.40}$\times$) \\
			FedNTD                                                & 87.66 (\textit{1.60}$\times$)                       & 47.01 (\textit{5.13}$\times$)          & 60.69 (\textit{1.68}$\times$)                           & 65.31 (\textit{1.71}$\times$)          & 68.10 (\textit{1.36}$\times$)          & 33.75 (\textit{1.48}$\times$)          & 57.66 (\textit{1.57}$\times$)          \\ \midrule
			FedLMD                                          & \textbf{88.61} (\textit{2.47}$\times$)              & \textbf{50.45} (\textit{5.26}$\times$) & \textbf{61.32} (\textit{1.77}$\times$)                  & \textbf{66.43} (\textit{2.04}$\times$) & \textbf{68.67} (\textit{1.39}$\times$)          & 33.72 (\textit{1.46}$\times$)          & 57.73 (\textit{1.57}$\times$)          \\ \bottomrule
		\end{tabular}
	}
\end{table*}

\subsection{Discussion}

\myPara{Model Architecture.}~We verify the applicability of the approach with different network architectures, and illustrate the performance with several typical architectures on CIFAR-10 ($\alpha=0.05$). As shown in~\tabref{tab:abl_model}, FedLMD works well with these network architectures. 

\begin{table}[H]
    \caption{The top-1 test accuracy (\%) under different networks.}
    \label{tab:abl_model}
    \centering
    \begin{tabular}{l|cccc}
        \toprule
        Method          & CNN   & MobileNet & ResNet-8   \\ \midrule
        FedAvg          & 33.02 & 27.84     & 30.96    & \\ \midrule
        FedCurv         & 39.88 & 14.10     & 28.10    & \\
        FedProx         & 42.74 & 30.83     & 31.32    & \\
        FedNTD          & 47.01 & 30.94     & 31.41    & \\ \midrule
        FedLMD    & \textbf{50.45} & \textbf{31.76}     & \textbf{34.60}    & \\ \bottomrule
    \end{tabular}
\end{table}

\begin{figure}[htbp]
    \centering
    \includegraphics[width=\linewidth]{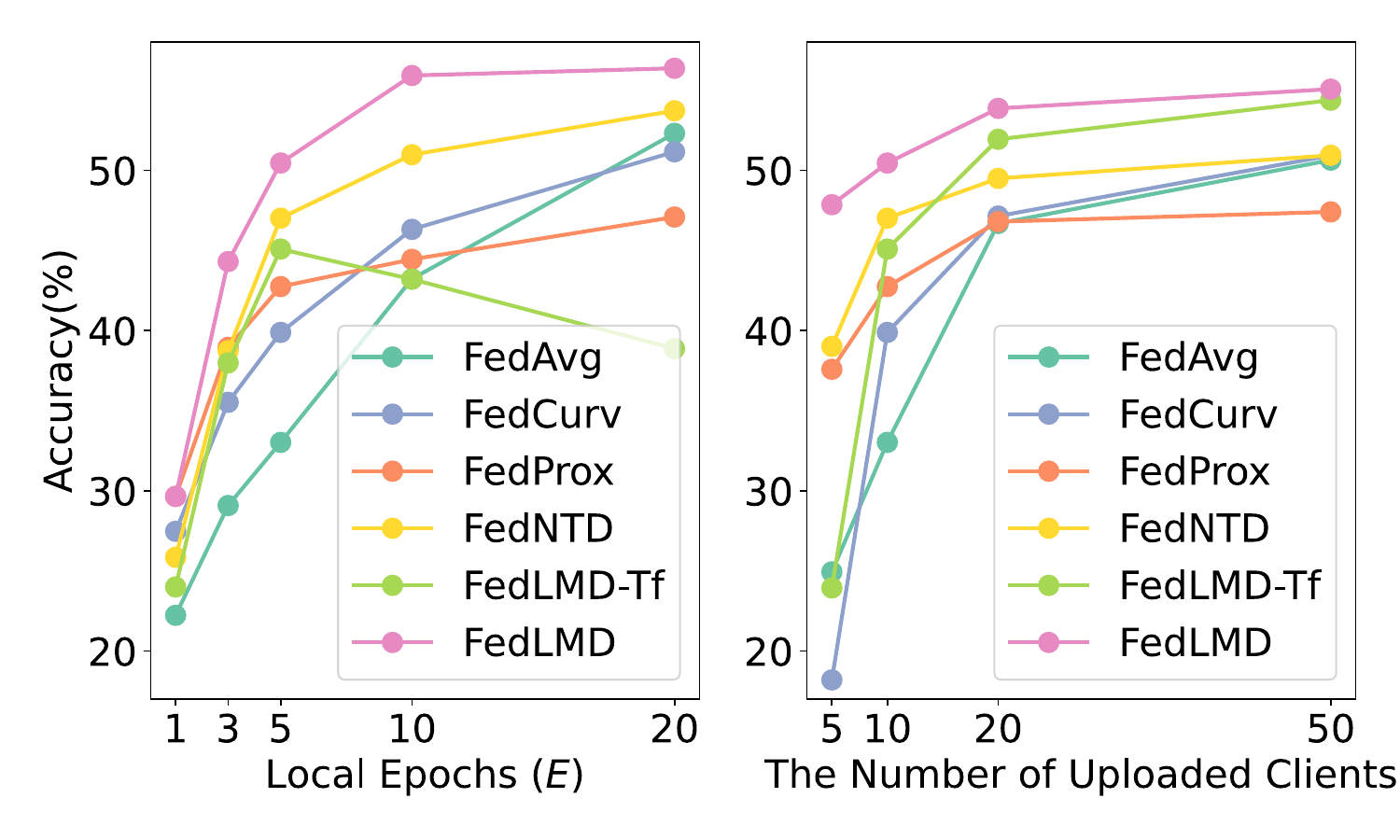}
    \caption{The top-1 test accuracy (\%) with different numbers of local epochs (Left) and the uploaded different number of client models (Right).} 
    \label{fig:epoch_client}
\end{figure}

\myPara{Local Epoch Number.}~We study the effect of local training epochs on accuracy, and report the results on the left of~\figref{fig:epoch_client}. As for FedLMD, the enhancement is stable and with excellent performance. While FedLMD-Tf does not perform as well as expected. When $E=10$, the performance of FedLMD-Tf starts to deteriorate (green line). It may indicate that the client-side model will rely too much on the teacher's performance as $E$ increases, and an unreliable teacher like a fixed distribution vector will limit the optimization of the client-side model with too many local epochs. It is worth pointing out that a larger $E$ will also increase the computational overhead of the client and not all scenarios are better with a larger $E$~\cite{fedavg}.

\myPara{Number of Uploaded Clients.}~Another point worth discussing in the FL is the number of uploaded clients per communication round. As shown in the right of~\figref{fig:epoch_client}, the optimal accuracy of each method increases with the number of participating clients. It can be found that FedLMD can achieve good results without having too many models for aggregation, which may be due to its ability to effectively preserve the knowledge of minority labels with a small number of clients aggregated. As for FedLMD-Tf, it performs similarly to FedAvg when the number of uploaded clients is low. While, when the number of clients increases to 20, FedLMD-Tf has surpassed SOTA baselines with additional computing resources (such as FedNTD). And when the number of clients is 50, it is already quite close to the teacher version (FedLMD). 

\myPara{Combination with Other FL Methods.}~We consider the combination of FedLMD and other FL methods for improvement. Here, we select two representative methods, FedProx~\cite{fedprox} and FedAvgM~\cite{fedavgm}. FedProx constrains the optimization of the local model from the perspective of model parameters. FedAvgM is based on the Adam optimization algorithm and incorporates a momentum parameter on top of FedAvg. By combining the previous global model parameters with the current aggregated global parameters, it updates the global parameters. As shown in~\tabref{tab:exp_othermethod}, the combination of FedLMD and FedAvgM performs better than FedLMD on CIFAR-10 ($\alpha=0.5$), which indicates that the combination of FedLMD and FedAvgM can be applied simultaneously when the label heterogeneity is not very high. Due to the fact that both FedLMD and FedProx are optimization methods with parameter constraints, their optimization trajectories may clash and compromise system performance.


\begin{table}[H]
    \centering
    \caption{The top-1 test accuracy (\%) under the combination of FedLMD and other federated learning methods.}
    \label{tab:exp_othermethod}
    \begin{tabular}{l|c|c|c|c}
        \toprule
        Method                 & FedLMD & +Prox & +AvgM & +Prox+AvgM  \\ \midrule
        Accuracy               & 68.67   & 65.42   &  \textbf{71.50} &  70.33    \\ \bottomrule
    \end{tabular}
\end{table}

\begin{figure}[!t]
    \centering
    \includegraphics[width=1.0\linewidth]{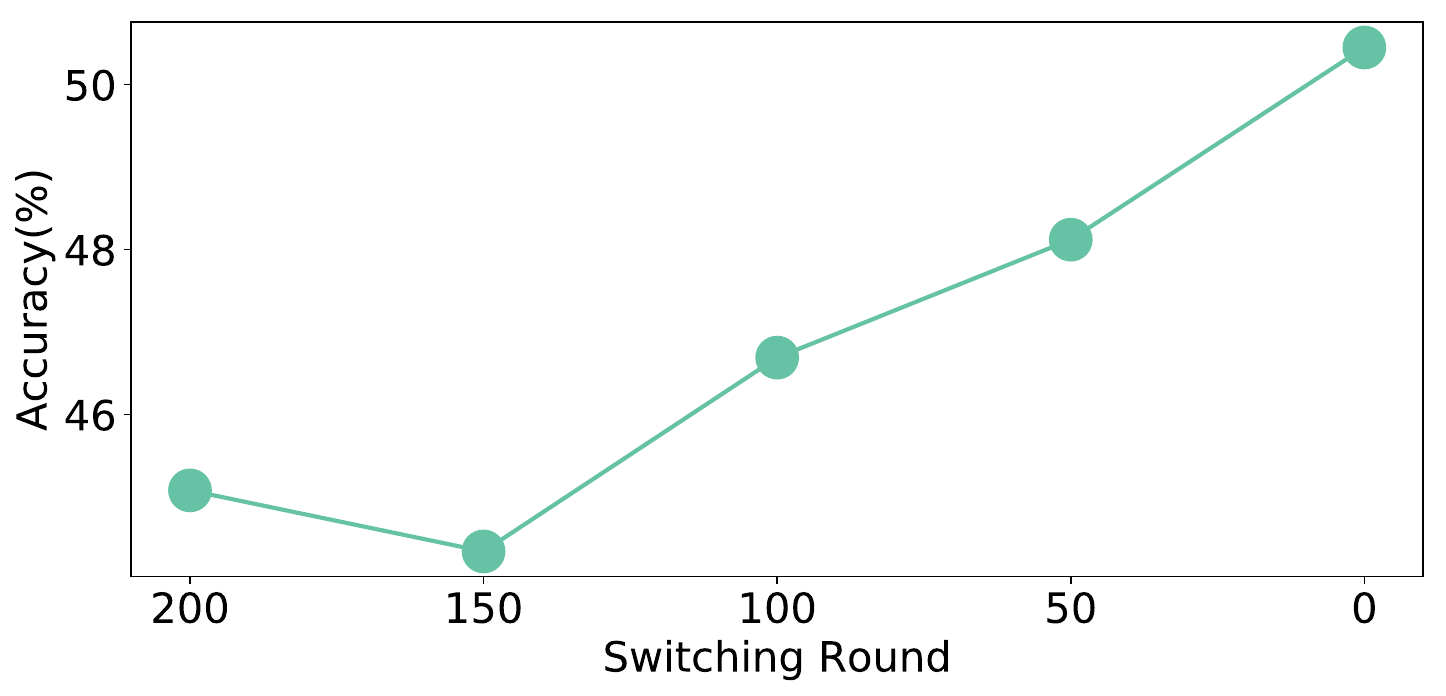}
    \caption{The top-1 test accuracy (\%) when switching from FedLMD-Tf to FedLMD after varying rounds.}
    \label{fig:method_mix}
\end{figure}

\myPara{Switching from FedLMD-Tf to FedLMD.}~As stated in~\secref{sec:lmd2tf}, the difference between FedLMD-Tf and FedLMD lies in whether the teacher is used or not. FedLMD-Tf is computationally efficient without a teacher, while FedLMD has a high performance with the teacher. We consider performing FedLMD-TF first and then switching to FedLMD later since the global model is not a good teacher in beginning.
As shown in~\figref{fig:method_mix}, we show that the optimization objective improves the performance of the method under different communication rounds of switching from FedLMD-TF to FedLMD on CIFAR-10 ($\alpha=0.05$). When the switching round is 200, the method becomes FedLMD-Tf, and when the switching round is 0, it is FedLMD. According to~\figref{fig:method_mix}, the performance can be improved by earlier turn switching, which is also accompanied by an increase in computational cost. With such improvements, we can select which round to switch according to actual training situation for balancing between performance and computation.

\subsection{Hyperparameters Analysis}
\begin{figure}[htbp]
    \centering
\includegraphics[width=\linewidth]{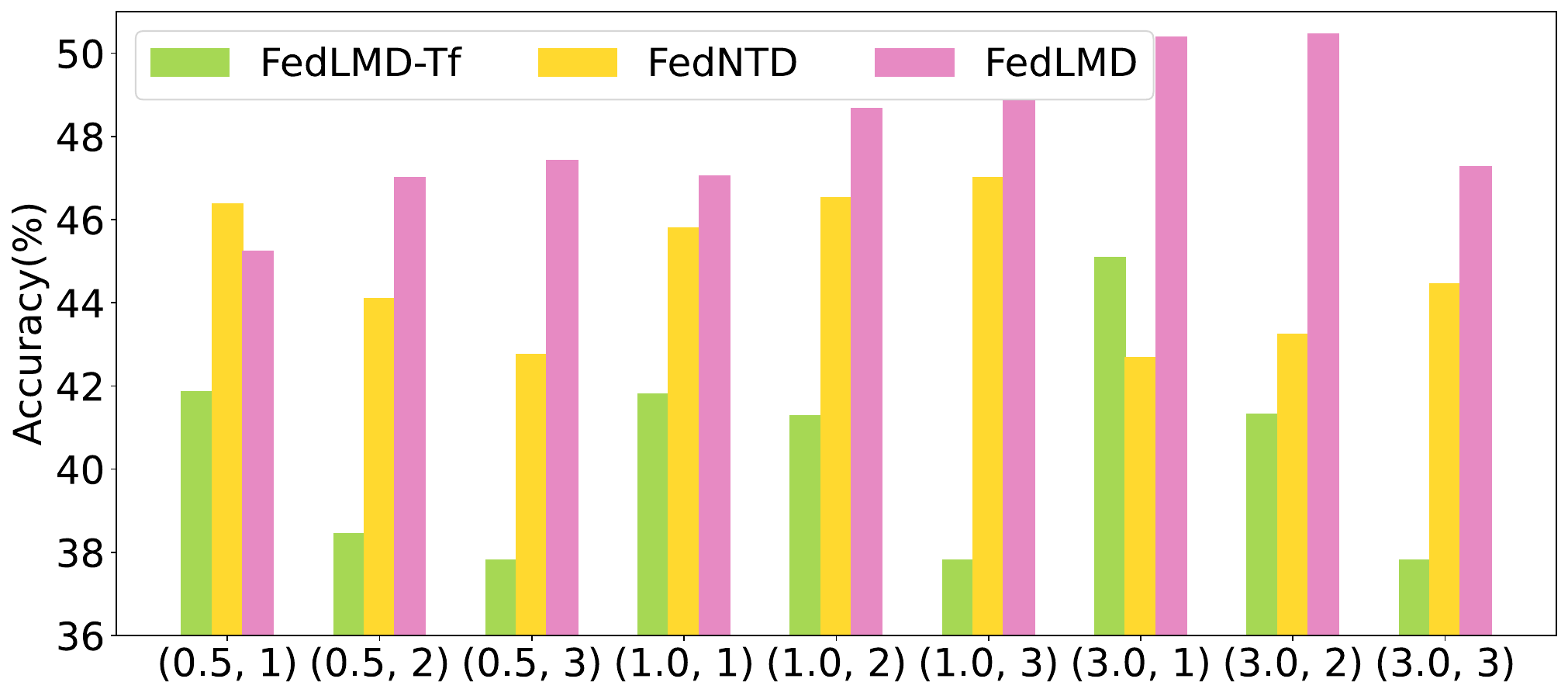}
    \caption{The top-1 test accuracy (\%) with different distillation methods under different hyperparameters ($\tau$, $\beta$) settings on CIFAR-10 ($\alpha=0.05$).} %
    \label{fig:exp_abl_hyperv}
\end{figure}

\figref{fig:exp_abl_hyperv} shows the performance of the proposed approach under different hyperparameters. FedLMD achieves excellent performance in most cases, which shows its robustness to the choice of hyperparameters. And for FedLMD-Tf, it suffers from severe performance degradation at higher $\beta$. This is mainly due to an unreliable teacher constraining the optimization of the local model. For the temperature $\tau$, a higher value leads to a better performance of FedLMD-Tf, which indicates that a smoother output of the local model is conducive to knowledge retention via teacher-free distillation.

\section{Conclusion}

In this paper, we propose FedLMD solve the challenge of label distribution skew in data heterogeneity, which achieves effective and stable FL by retaining knowledge of minority labels. It does not require additional parameters to be uploaded, and thus does not carry additional communication overhead and privacy risk. Our experimental results show that FedLMD is more effective compared to previous methods. Further, considering the limited computational resources on the client-side, we improve it to a teacher-free version. It achieves excellent performance without additional computation. In future work, we will focus on how to apply in larger-scale application scenarios and the optimization solution for other data heterogeneous cases, like the rare labels in the all clients.


\bibliographystyle{main}
\bibliography{main}


\begin{thebibliography}{52}


\ifx \showCODEN    \undefined \def \showCODEN     #1{\unskip}     \fi
\ifx \showDOI      \undefined \def \showDOI       #1{#1}\fi
\ifx \showISBNx    \undefined \def \showISBNx     #1{\unskip}     \fi
\ifx \showISBNxiii \undefined \def \showISBNxiii  #1{\unskip}     \fi
\ifx \showISSN     \undefined \def \showISSN      #1{\unskip}     \fi
\ifx \showLCCN     \undefined \def \showLCCN      #1{\unskip}     \fi
\ifx \shownote     \undefined \def \shownote      #1{#1}          \fi
\ifx \showarticletitle \undefined \def \showarticletitle #1{#1}   \fi
\ifx \showURL      \undefined \def \showURL       {\relax}        \fi
\providecommand\bibfield[2]{#2}
\providecommand\bibinfo[2]{#2}
\providecommand\natexlab[1]{#1}
\providecommand\showeprint[2][]{arXiv:#2}

\bibitem[Acar et~al\mbox{.}(2021)]%
        {feddyn}
\bibfield{author}{\bibinfo{person}{Durmus Alp~Emre Acar}, \bibinfo{person}{Yue Zhao}, \bibinfo{person}{Ramon Matas}, \bibinfo{person}{Matthew Mattina}, \bibinfo{person}{Paul Whatmough}, {and} \bibinfo{person}{Venkatesh Saligrama}.} \bibinfo{year}{2021}\natexlab{}.
\newblock \showarticletitle{Federated Learning Based on Dynamic Regularization}. In \bibinfo{booktitle}{\emph{ICLR}}.
\newblock


\bibitem[Bukaty(2019)]%
        {10.2307/j.ctvjghvnn}
\bibfield{author}{\bibinfo{person}{Preston Bukaty}.} \bibinfo{year}{2019}\natexlab{}.
\newblock \bibinfo{booktitle}{\emph{The California Consumer Privacy Act (CCPA): An Implementation Guide}}.
\newblock \bibinfo{publisher}{IT Governance Publishing}.
\newblock


\bibitem[Chen and Chao(2021)]%
        {chen_fedbe_2021}
\bibfield{author}{\bibinfo{person}{Hongyou Chen} {and} \bibinfo{person}{Weilun Chao}.} \bibinfo{year}{2021}\natexlab{}.
\newblock \showarticletitle{FedBE: Making Bayesian Model Ensemble Applicable to Federated Learning}. In \bibinfo{booktitle}{\emph{ICLR}}.
\newblock


\bibitem[Cho et~al\mbox{.}(2022)]%
        {cho2022heterogeneous}
\bibfield{author}{\bibinfo{person}{Yae~Jee Cho}, \bibinfo{person}{Andre Manoel}, \bibinfo{person}{Gauri Joshi}, \bibinfo{person}{Robert Sim}, {and} \bibinfo{person}{Dimitrios Dimitriadis}.} \bibinfo{year}{2022}\natexlab{}.
\newblock \showarticletitle{Heterogeneous Ensemble Knowledge Transfer for Training Large Models in Federated Learning}. In \bibinfo{booktitle}{\emph{IJCAI}}. \bibinfo{pages}{2881--2887}.
\newblock


\bibitem[Darlow et~al\mbox{.}(2018)]%
        {darlow2018cinic}
\bibfield{author}{\bibinfo{person}{Luke~N Darlow}, \bibinfo{person}{Elliot~J Crowley}, \bibinfo{person}{Antreas Antoniou}, {and} \bibinfo{person}{Amos~J Storkey}.} \bibinfo{year}{2018}\natexlab{}.
\newblock \showarticletitle{Cinic-10 is not imagenet or cifar-10}.
\newblock \bibinfo{journal}{\emph{arXiv:1810.03505}} (\bibinfo{year}{2018}).
\newblock


\bibitem[Gong et~al\mbox{.}(2021)]%
        {gong2021iccv}
\bibfield{author}{\bibinfo{person}{Xuan Gong}, \bibinfo{person}{Abhishek Sharma}, \bibinfo{person}{Srikrishna Karanam}, \bibinfo{person}{Ziyan Wu}, \bibinfo{person}{Terrence Chen}, \bibinfo{person}{David~S. Doermann}, {and} \bibinfo{person}{Arun Innanje}.} \bibinfo{year}{2021}\natexlab{}.
\newblock \showarticletitle{Ensemble Attention Distillation for Privacy-Preserving Federated Learning}. In \bibinfo{booktitle}{\emph{ICCV}}. \bibinfo{pages}{15056--15066}.
\newblock


\bibitem[Han et~al\mbox{.}(2020)]%
        {00030YYXTS20}
\bibfield{author}{\bibinfo{person}{Bo Han}, \bibinfo{person}{Gang Niu}, \bibinfo{person}{Xingrui Yu}, \bibinfo{person}{Quanming Yao}, \bibinfo{person}{Miao Xu}, \bibinfo{person}{Ivor~W. Tsang}, {and} \bibinfo{person}{Masashi Sugiyama}.} \bibinfo{year}{2020}\natexlab{}.
\newblock \showarticletitle{{SIGUA:} Forgetting May Make Learning with Noisy Labels More Robust}. In \bibinfo{booktitle}{\emph{ICML}}. \bibinfo{pages}{4006--4016}.
\newblock


\bibitem[He et~al\mbox{.}(2022a)]%
        {he2022learning}
\bibfield{author}{\bibinfo{person}{Yuting He}, \bibinfo{person}{Yiqiang Chen}, \bibinfo{person}{XiaoDong Yang}, \bibinfo{person}{Hanchao Yu}, \bibinfo{person}{Yi-Hua Huang}, {and} \bibinfo{person}{Yang Gu}.} \bibinfo{year}{2022}\natexlab{a}.
\newblock \showarticletitle{Learning Critically: Selective Self-Distillation in Federated Learning on Non-IID Data}.
\newblock \bibinfo{journal}{\emph{IEEE Trans. Big Data}} (\bibinfo{year}{2022}).
\newblock


\bibitem[He et~al\mbox{.}(2022b)]%
        {he2022class}
\bibfield{author}{\bibinfo{person}{Yuting He}, \bibinfo{person}{Yiqiang Chen}, \bibinfo{person}{Xiaodong Yang}, \bibinfo{person}{Yingwei Zhang}, {and} \bibinfo{person}{Bixiao Zeng}.} \bibinfo{year}{2022}\natexlab{b}.
\newblock \showarticletitle{Class-Wise Adaptive Self Distillation for Federated Learning on Non-IID Data (Student Abstract)}. In \bibinfo{booktitle}{\emph{AAAI}}. \bibinfo{pages}{12967--12968}.
\newblock


\bibitem[Hinton et~al\mbox{.}(2015)]%
        {hintonDistillingKnowledgeNeural2015}
\bibfield{author}{\bibinfo{person}{Geoffrey Hinton}, \bibinfo{person}{Oriol Vinyals}, {and} \bibinfo{person}{Jeff Dean}.} \bibinfo{year}{2015}\natexlab{}.
\newblock \showarticletitle{Distilling the {{Knowledge}} in a {{Neural Network}}}. In \bibinfo{booktitle}{\emph{NeurIPS Workshop}}.
\newblock


\bibitem[Hsu et~al\mbox{.}(2019)]%
        {DBLP:journals/corr/abs-1909-06335}
\bibfield{author}{\bibinfo{person}{Tzu{-}Ming~Harry Hsu}, \bibinfo{person}{Hang Qi}, {and} \bibinfo{person}{Matthew Brown}.} \bibinfo{year}{2019}\natexlab{}.
\newblock \showarticletitle{Measuring the Effects of Non-Identical Data Distribution for Federated Visual Classification}.
\newblock \bibinfo{journal}{\emph{arXiv:1909.06335}} (\bibinfo{year}{2019}).
\newblock


\bibitem[Hu et~al\mbox{.}(2022)]%
        {10.1145/3501814}
\bibfield{author}{\bibinfo{person}{Ziheng Hu}, \bibinfo{person}{Hongtao Xie}, \bibinfo{person}{Lingyun Yu}, \bibinfo{person}{Xingyu Gao}, \bibinfo{person}{Zhihua Shang}, {and} \bibinfo{person}{Yongdong Zhang}.} \bibinfo{year}{2022}\natexlab{}.
\newblock \showarticletitle{Dynamic-Aware Federated Learning for Face Forgery Video Detection}.
\newblock \bibinfo{journal}{\emph{ACM Trans. Intell. Syst. Technol.}} \bibinfo{volume}{13}, \bibinfo{number}{4} (\bibinfo{year}{2022}).
\newblock
\showISSN{2157-6904}


\bibitem[Jeong et~al\mbox{.}(2018)]%
        {jeong2018communication}
\bibfield{author}{\bibinfo{person}{Eunjeong Jeong}, \bibinfo{person}{Seungeun Oh}, \bibinfo{person}{Hyesung Kim}, {et~al\mbox{.}}} \bibinfo{year}{2018}\natexlab{}.
\newblock \showarticletitle{Communication-Efficient On-Device Machine Learning: Federated Distillation and Augmentation under Non-IID Private Data}. In \bibinfo{booktitle}{\emph{NeurIPS Workshop}}.
\newblock


\bibitem[Kairouz et~al\mbox{.}(2021)]%
        {kairouz2019advances}
\bibfield{author}{\bibinfo{person}{Peter Kairouz}, \bibinfo{person}{H~Brendan McMahan}, \bibinfo{person}{Brendan Avent}, {et~al\mbox{.}}} \bibinfo{year}{2021}\natexlab{}.
\newblock \showarticletitle{Advances and Open Problems in Federated Learning}.
\newblock \bibinfo{journal}{\emph{Found. Trends Mach. Learn.}} \bibinfo{volume}{14}, \bibinfo{number}{1-2} (\bibinfo{year}{2021}), \bibinfo{pages}{1--210}.
\newblock


\bibitem[Karimireddy et~al\mbox{.}(2020)]%
        {scaffold}
\bibfield{author}{\bibinfo{person}{Sai~Praneeth Karimireddy}, \bibinfo{person}{Satyen Kale}, \bibinfo{person}{Mehryar Mohri}, \bibinfo{person}{Sashank~J. Reddi}, \bibinfo{person}{Sebastian~U. Stich}, {and} \bibinfo{person}{Ananda~Theertha Suresh}.} \bibinfo{year}{2020}\natexlab{}.
\newblock \showarticletitle{{SCAFFOLD:} Stochastic Controlled Averaging for Federated Learning}. In \bibinfo{booktitle}{\emph{ICML}}. \bibinfo{pages}{5132--5143}.
\newblock


\bibitem[Kim et~al\mbox{.}(2019)]%
        {KimYYK19}
\bibfield{author}{\bibinfo{person}{Youngdong Kim}, \bibinfo{person}{Junho Yim}, \bibinfo{person}{Juseung Yun}, {and} \bibinfo{person}{Junmo Kim}.} \bibinfo{year}{2019}\natexlab{}.
\newblock \showarticletitle{{NLNL:} Negative Learning for Noisy Labels}. In \bibinfo{booktitle}{\emph{ICCV}}. \bibinfo{pages}{101--110}.
\newblock


\bibitem[Krizhevsky et~al\mbox{.}(2009)]%
        {krizhevsky2009learning}
\bibfield{author}{\bibinfo{person}{Alex Krizhevsky}, \bibinfo{person}{Geoffrey Hinton}, {et~al\mbox{.}}} \bibinfo{year}{2009}\natexlab{}.
\newblock \showarticletitle{Learning multiple layers of features from tiny images}.
\newblock \bibinfo{journal}{\emph{Master's thesis, Department of Computer Science, University of Toronto}} (\bibinfo{year}{2009}).
\newblock


\bibitem[Lecun et~al\mbox{.}(1998)]%
        {726791}
\bibfield{author}{\bibinfo{person}{Y. Lecun}, \bibinfo{person}{L. Bottou}, \bibinfo{person}{Y. Bengio}, {and} \bibinfo{person}{P. Haffner}.} \bibinfo{year}{1998}\natexlab{}.
\newblock \showarticletitle{Gradient-based learning applied to document recognition}.
\newblock \bibinfo{journal}{\emph{Proc. IEEE}} \bibinfo{volume}{86}, \bibinfo{number}{11} (\bibinfo{year}{1998}), \bibinfo{pages}{2278--2324}.
\newblock


\bibitem[Lee et~al\mbox{.}(2022)]%
        {fedntd}
\bibfield{author}{\bibinfo{person}{Gihun Lee}, \bibinfo{person}{Minchan Jeong}, \bibinfo{person}{Yongjin Shin}, \bibinfo{person}{Sangmin Bae}, {and} \bibinfo{person}{Se-Young Yun}.} \bibinfo{year}{2022}\natexlab{}.
\newblock \showarticletitle{Preservation of Global Knowledge by Not-True Distillation in Federated Learning}. In \bibinfo{booktitle}{\emph{NeurIPS}}. \bibinfo{pages}{38461--38474}.
\newblock


\bibitem[Li et~al\mbox{.}(2021)]%
        {moon}
\bibfield{author}{\bibinfo{person}{Qinbin Li}, \bibinfo{person}{Bingsheng He}, {and} \bibinfo{person}{Dawn Song}.} \bibinfo{year}{2021}\natexlab{}.
\newblock \showarticletitle{Model-Contrastive Federated Learning}. In \bibinfo{booktitle}{\emph{CVPR}}. \bibinfo{pages}{10713--10722}.
\newblock


\bibitem[Li et~al\mbox{.}(2020c)]%
        {li2020invisiblefl}
\bibfield{author}{\bibinfo{person}{Qiushi Li}, \bibinfo{person}{Wenwu Zhu}, \bibinfo{person}{Chao Wu}, \bibinfo{person}{Xinglin Pan}, \bibinfo{person}{Fan Yang}, \bibinfo{person}{Yuezhi Zhou}, {and} \bibinfo{person}{Yaoxue Zhang}.} \bibinfo{year}{2020}\natexlab{c}.
\newblock \showarticletitle{InvisibleFL: federated learning over non-informative intermediate updates against multimedia privacy leakages}. In \bibinfo{booktitle}{\emph{ACM Multimedia}}. \bibinfo{pages}{753--762}.
\newblock


\bibitem[Li et~al\mbox{.}(2020a)]%
        {li2020federated}
\bibfield{author}{\bibinfo{person}{Tian Li}, \bibinfo{person}{Anit~Kumar Sahu}, \bibinfo{person}{Ameet Talwalkar}, {and} \bibinfo{person}{Virginia Smith}.} \bibinfo{year}{2020}\natexlab{a}.
\newblock \showarticletitle{Federated Learning: Challenges, Methods, and Future Directions}.
\newblock \bibinfo{journal}{\emph{IEEE Signal Processing Magazine}} \bibinfo{volume}{37}, \bibinfo{number}{3} (\bibinfo{year}{2020}), \bibinfo{pages}{50--60}.
\newblock


\bibitem[Li et~al\mbox{.}(2020b)]%
        {fedprox}
\bibfield{author}{\bibinfo{person}{Tian Li}, \bibinfo{person}{Anit~Kumar Sahu}, \bibinfo{person}{Manzil Zaheer}, \bibinfo{person}{Maziar Sanjabi}, \bibinfo{person}{Ameet Talwalkar}, {and} \bibinfo{person}{Virginia Smith}.} \bibinfo{year}{2020}\natexlab{b}.
\newblock \showarticletitle{Federated Optimization in Heterogeneous Networks}. In \bibinfo{booktitle}{\emph{MLSys}}. \bibinfo{pages}{429--450}.
\newblock


\bibitem[Li et~al\mbox{.}(2019)]%
        {li2019privacy}
\bibfield{author}{\bibinfo{person}{Wenqi Li}, \bibinfo{person}{Fausto Milletar{\`\i}}, \bibinfo{person}{Daguang Xu}, \bibinfo{person}{Nicola Rieke}, \bibinfo{person}{Jonny Hancox}, \bibinfo{person}{Wentao Zhu}, \bibinfo{person}{Maximilian Baust}, \bibinfo{person}{Yan Cheng}, \bibinfo{person}{S{\'e}bastien Ourselin}, \bibinfo{person}{M~Jorge Cardoso}, {et~al\mbox{.}}} \bibinfo{year}{2019}\natexlab{}.
\newblock \showarticletitle{Privacy-Preserving Federated Brain Tumour Segmentation}. In \bibinfo{booktitle}{\emph{MLMI}}. \bibinfo{pages}{133--141}.
\newblock


\bibitem[Li and Zhan(2021)]%
        {fedrs}
\bibfield{author}{\bibinfo{person}{Xin-Chun Li} {and} \bibinfo{person}{De-Chuan Zhan}.} \bibinfo{year}{2021}\natexlab{}.
\newblock \showarticletitle{FedRS: Federated Learning with Restricted Softmax for Label Distribution Non-IID Data}. In \bibinfo{booktitle}{\emph{SIGKDD}}. \bibinfo{pages}{995–1005}.
\newblock
\showISBNx{9781450383325}


\bibitem[Lin et~al\mbox{.}(2020)]%
        {lin_ensemble_2020}
\bibfield{author}{\bibinfo{person}{Tao Lin}, \bibinfo{person}{Lingjing Kong}, \bibinfo{person}{Sebastian~U Stich}, {and} \bibinfo{person}{Martin Jaggi}.} \bibinfo{year}{2020}\natexlab{}.
\newblock \showarticletitle{Ensemble Distillation for Robust Model Fusion in Federated Learning}. In \bibinfo{booktitle}{\emph{NeurIPS}}. \bibinfo{pages}{2351--2363}.
\newblock


\bibitem[Liu et~al\mbox{.}(2021)]%
        {liu2021feddg}
\bibfield{author}{\bibinfo{person}{Quande Liu}, \bibinfo{person}{Cheng Chen}, \bibinfo{person}{Jing Qin}, \bibinfo{person}{Qi Dou}, {and} \bibinfo{person}{Pheng-Ann Heng}.} \bibinfo{year}{2021}\natexlab{}.
\newblock \showarticletitle{FedDG: Federated Domain Generalization on Medical Image Segmentation via Episodic Learning in Continuous Frequency Space}.
\newblock \bibinfo{journal}{\emph{CVPR}} (\bibinfo{year}{2021}).
\newblock


\bibitem[McMahan et~al\mbox{.}(2017)]%
        {fedavg}
\bibfield{author}{\bibinfo{person}{Brendan McMahan}, \bibinfo{person}{Eider Moore}, \bibinfo{person}{Daniel Ramage}, \bibinfo{person}{Seth Hampson}, {and} \bibinfo{person}{Blaise~Aguera y Arcas}.} \bibinfo{year}{2017}\natexlab{}.
\newblock \showarticletitle{Communication-Efficient Learning of Deep Networks from Decentralized Data}. In \bibinfo{booktitle}{\emph{AISTATS}}, Vol.~\bibinfo{volume}{54}. \bibinfo{pages}{1273--1282}.
\newblock


\bibitem[Mothukuri et~al\mbox{.}(2021)]%
        {mothukuri2021survey}
\bibfield{author}{\bibinfo{person}{Viraaji Mothukuri}, \bibinfo{person}{Reza~M Parizi}, \bibinfo{person}{Seyedamin Pouriyeh}, \bibinfo{person}{Yan Huang}, \bibinfo{person}{Ali Dehghantanha}, {and} \bibinfo{person}{Gautam Srivastava}.} \bibinfo{year}{2021}\natexlab{}.
\newblock \showarticletitle{A survey on security and privacy of federated learning}.
\newblock \bibinfo{journal}{\emph{Future Generation Computer Systems}}  \bibinfo{volume}{115} (\bibinfo{year}{2021}), \bibinfo{pages}{619--640}.
\newblock


\bibitem[Nguyen et~al\mbox{.}(2023)]%
        {DBLP:journals/csur/NguyenPPDSLDH23}
\bibfield{author}{\bibinfo{person}{Dinh~C. Nguyen}, \bibinfo{person}{Quoc{-}Viet Pham}, \bibinfo{person}{Pubudu~N. Pathirana}, \bibinfo{person}{Ming Ding}, \bibinfo{person}{Aruna Seneviratne}, \bibinfo{person}{Zihuai Lin}, \bibinfo{person}{Octavia~A. Dobre}, {and} \bibinfo{person}{Won{-}Joo Hwang}.} \bibinfo{year}{2023}\natexlab{}.
\newblock \showarticletitle{Federated Learning for Smart Healthcare: {A} Survey}.
\newblock \bibinfo{journal}{\emph{{ACM} Comput. Surv.}} \bibinfo{volume}{55}, \bibinfo{number}{3} (\bibinfo{year}{2023}), \bibinfo{pages}{60:1--60:37}.
\newblock


\bibitem[Pan and Sun(2021)]%
        {DBLP:journals/corr/abs-2107-00051}
\bibfield{author}{\bibinfo{person}{Wanning Pan} {and} \bibinfo{person}{Lichao Sun}.} \bibinfo{year}{2021}\natexlab{}.
\newblock \showarticletitle{Local-Global Knowledge Distillation in Heterogeneous Federated Learning with Non-IID Data}.
\newblock \bibinfo{journal}{\emph{arXiv:2107.00051}} (\bibinfo{year}{2021}).
\newblock


\bibitem[Philipp(2016)]%
        {philipp2016gdpr}
\bibfield{author}{\bibinfo{person}{Albrecht~Jan Philipp}.} \bibinfo{year}{2016}\natexlab{}.
\newblock \showarticletitle{How the GDPR will change the world}.
\newblock \bibinfo{journal}{\emph{European Data Protection Law Review}} \bibinfo{volume}{2}, \bibinfo{number}{3} (\bibinfo{year}{2016}), \bibinfo{pages}{287}.
\newblock


\bibitem[Qi et~al\mbox{.}(2022a)]%
        {10.1145/3503161.3548278}
\bibfield{author}{\bibinfo{person}{Fan Qi}, \bibinfo{person}{Zixin Zhang}, \bibinfo{person}{Xianshan Yang}, \bibinfo{person}{Huaiwen Zhang}, {and} \bibinfo{person}{Changsheng Xu}.} \bibinfo{year}{2022}\natexlab{a}.
\newblock \showarticletitle{Feeling Without Sharing: A Federated Video Emotion Recognition Framework Via Privacy-Agnostic Hybrid Aggregation}. In \bibinfo{booktitle}{\emph{ACM Multimedia}} (Lisboa, Portugal) \emph{(\bibinfo{series}{MM '22})}. \bibinfo{pages}{151–160}.
\newblock


\bibitem[Qi et~al\mbox{.}(2022b)]%
        {DBLP:conf/mm/QiZYZX22}
\bibfield{author}{\bibinfo{person}{Fan Qi}, \bibinfo{person}{Zixin Zhang}, \bibinfo{person}{Xianshan Yang}, \bibinfo{person}{Huaiwen Zhang}, {and} \bibinfo{person}{Changsheng Xu}.} \bibinfo{year}{2022}\natexlab{b}.
\newblock \showarticletitle{Feeling Without Sharing: {A} Federated Video Emotion Recognition Framework Via Privacy-Agnostic Hybrid Aggregation}. In \bibinfo{booktitle}{\emph{ACM Multimedia}}. \bibinfo{pages}{151--160}.
\newblock


\bibitem[Remedios et~al\mbox{.}(2020)]%
        {fedavgm}
\bibfield{author}{\bibinfo{person}{Samuel~W. Remedios}, \bibinfo{person}{John~A. Butman}, \bibinfo{person}{Bennett~A. Landman}, {and} \bibinfo{person}{Dzung~L. Pham}.} \bibinfo{year}{2020}\natexlab{}.
\newblock \showarticletitle{Federated Gradient Averaging for Multi-Site Training with Momentum-Based Optimizers}. In \bibinfo{booktitle}{\emph{MICCAI Workshop}} \emph{(\bibinfo{series}{Lecture Notes in Computer Science}, Vol.~\bibinfo{volume}{12444})}. \bibinfo{pages}{170--180}.
\newblock


\bibitem[Sattler et~al\mbox{.}(2021)]%
        {sattler2021tnnls}
\bibfield{author}{\bibinfo{person}{Felix Sattler}, \bibinfo{person}{Tim Korjakow}, \bibinfo{person}{Roman Rischke}, {and} \bibinfo{person}{Wojciech Samek}.} \bibinfo{year}{2021}\natexlab{}.
\newblock \showarticletitle{FedAUX: Leveraging Unlabeled Auxiliary Data in Federated Learning}.
\newblock \bibinfo{journal}{\emph{IEEE TNNLS}} (\bibinfo{year}{2021}), \bibinfo{pages}{1--13}.
\newblock


\bibitem[Shahid et~al\mbox{.}(2021)]%
        {shahid2021communication}
\bibfield{author}{\bibinfo{person}{Osama Shahid}, \bibinfo{person}{Seyedamin Pouriyeh}, \bibinfo{person}{Reza~M Parizi}, \bibinfo{person}{Quan~Z Sheng}, \bibinfo{person}{Gautam Srivastava}, {and} \bibinfo{person}{Liang Zhao}.} \bibinfo{year}{2021}\natexlab{}.
\newblock \showarticletitle{Communication Efficiency in Federated Learning: Achievements and Challenges}.
\newblock \bibinfo{journal}{\emph{arXiv:2107.10996}} (\bibinfo{year}{2021}).
\newblock


\bibitem[Shi et~al\mbox{.}(2021)]%
        {DBLP:journals/corr/abs-2109-14611}
\bibfield{author}{\bibinfo{person}{Haizhou Shi}, \bibinfo{person}{Youcai Zhang}, \bibinfo{person}{Zijin Shen}, \bibinfo{person}{Siliang Tang}, \bibinfo{person}{Yaqian Li}, \bibinfo{person}{Yandong Guo}, {and} \bibinfo{person}{Yueting Zhuang}.} \bibinfo{year}{2021}\natexlab{}.
\newblock \showarticletitle{Federated Self-Supervised Contrastive Learning via Ensemble Similarity Distillation}.
\newblock \bibinfo{journal}{\emph{arXiv:2109.14611}} (\bibinfo{year}{2021}).
\newblock


\bibitem[Shoham et~al\mbox{.}(2019)]%
        {fedcurv}
\bibfield{author}{\bibinfo{person}{Neta Shoham}, \bibinfo{person}{Tomer Avidor}, \bibinfo{person}{Aviv Keren}, \bibinfo{person}{Nadav Israel}, \bibinfo{person}{Daniel Benditkis}, \bibinfo{person}{Liron Mor-Yosef}, {and} \bibinfo{person}{Itai Zeitak}.} \bibinfo{year}{2019}\natexlab{}.
\newblock \showarticletitle{Overcoming forgetting in federated learning on non-iid data}.
\newblock \bibinfo{journal}{\emph{arXiv:1910.07796}} (\bibinfo{year}{2019}).
\newblock


\bibitem[Sui et~al\mbox{.}(2020)]%
        {sui2020emnlp}
\bibfield{author}{\bibinfo{person}{Dianbo Sui}, \bibinfo{person}{Yubo Chen}, \bibinfo{person}{Jun Zhao}, \bibinfo{person}{Yantao Jia}, \bibinfo{person}{Yuantao Xie}, {and} \bibinfo{person}{Weijian Sun}.} \bibinfo{year}{2020}\natexlab{}.
\newblock \showarticletitle{FedED: Federated Learning via Ensemble Distillation for Medical Relation Extraction}. In \bibinfo{booktitle}{\emph{EMNLP}}. \bibinfo{pages}{2118--2128}.
\newblock


\bibitem[Taya et~al\mbox{.}(2021)]%
        {DBLP:journals/corr/abs-2104-00352}
\bibfield{author}{\bibinfo{person}{Akihito Taya}, \bibinfo{person}{Takayuki Nishio}, \bibinfo{person}{Masahiro Morikura}, {and} \bibinfo{person}{Koji Yamamoto}.} \bibinfo{year}{2021}\natexlab{}.
\newblock \showarticletitle{Decentralized and Model-Free Federated Learning: Consensus-Based Distillation in Function Space}.
\newblock \bibinfo{journal}{\emph{arXiv:2104.00352}} (\bibinfo{year}{2021}).
\newblock


\bibitem[Wang et~al\mbox{.}(2020b)]%
        {wang_federated_2020}
\bibfield{author}{\bibinfo{person}{Hongyi Wang}, \bibinfo{person}{Mikhail Yurochkin}, \bibinfo{person}{Yuekai Sun}, {et~al\mbox{.}}} \bibinfo{year}{2020}\natexlab{b}.
\newblock \showarticletitle{Federated Learning with Matched Averaging}. In \bibinfo{booktitle}{\emph{ICLR}}.
\newblock


\bibitem[Wang et~al\mbox{.}(2020a)]%
        {fednova}
\bibfield{author}{\bibinfo{person}{Jianyu Wang}, \bibinfo{person}{Qinghua Liu}, \bibinfo{person}{Hao Liang}, \bibinfo{person}{Gauri Joshi}, {and} \bibinfo{person}{H~Vincent Poor}.} \bibinfo{year}{2020}\natexlab{a}.
\newblock \showarticletitle{Tackling the objective inconsistency problem in heterogeneous federated optimization}. In \bibinfo{booktitle}{\emph{NeurIPS}}. \bibinfo{pages}{7611--7623}.
\newblock


\bibitem[Wu et~al\mbox{.}(2021)]%
        {DBLP:journals/tpds/WuYW21}
\bibfield{author}{\bibinfo{person}{Xueyu Wu}, \bibinfo{person}{Xin Yao}, {and} \bibinfo{person}{Cho{-}Li Wang}.} \bibinfo{year}{2021}\natexlab{}.
\newblock \showarticletitle{FedSCR: Structure-Based Communication Reduction for Federated Learning}.
\newblock \bibinfo{journal}{\emph{{IEEE} Trans. Parallel Distributed Syst.}} \bibinfo{volume}{32}, \bibinfo{number}{7} (\bibinfo{year}{2021}), \bibinfo{pages}{1565--1577}.
\newblock


\bibitem[Xu et~al\mbox{.}(2022)]%
        {xu2022acceleration}
\bibfield{author}{\bibinfo{person}{Chencheng Xu}, \bibinfo{person}{Zhiwei Hong}, \bibinfo{person}{Minlie Huang}, {and} \bibinfo{person}{Tao Jiang}.} \bibinfo{year}{2022}\natexlab{}.
\newblock \showarticletitle{Acceleration of Federated Learning with Alleviated Forgetting in Local Training}.
\newblock \bibinfo{journal}{\emph{arXiv:2203.02645}} (\bibinfo{year}{2022}).
\newblock


\bibitem[Yao et~al\mbox{.}(2021)]%
        {fedgkd}
\bibfield{author}{\bibinfo{person}{Dezhong Yao}, \bibinfo{person}{Wanning Pan}, \bibinfo{person}{Yutong Dai}, \bibinfo{person}{Yao Wan}, \bibinfo{person}{Xiaofeng Ding}, \bibinfo{person}{Hai Jin}, \bibinfo{person}{Zheng Xu}, {and} \bibinfo{person}{Lichao Sun}.} \bibinfo{year}{2021}\natexlab{}.
\newblock \showarticletitle{Local-Global Knowledge Distillation in Heterogeneous Federated Learning with Non-IID Data}.
\newblock \bibinfo{journal}{\emph{arXiv:2107.00051}} (\bibinfo{year}{2021}).
\newblock


\bibitem[Yu et~al\mbox{.}(2020)]%
        {DBLP:conf/icml/YuRMK20}
\bibfield{author}{\bibinfo{person}{Felix~X. Yu}, \bibinfo{person}{Ankit~Singh Rawat}, \bibinfo{person}{Aditya~Krishna Menon}, {and} \bibinfo{person}{Sanjiv Kumar}.} \bibinfo{year}{2020}\natexlab{}.
\newblock \showarticletitle{Federated Learning with Only Positive Labels}. In \bibinfo{booktitle}{\emph{ICML}}. \bibinfo{pages}{10946--10956}.
\newblock


\bibitem[Yuan et~al\mbox{.}(2020)]%
        {tfkd}
\bibfield{author}{\bibinfo{person}{Li Yuan}, \bibinfo{person}{Francis E.~H. Tay}, \bibinfo{person}{Guilin Li}, \bibinfo{person}{Tao Wang}, {and} \bibinfo{person}{Jiashi Feng}.} \bibinfo{year}{2020}\natexlab{}.
\newblock \showarticletitle{Revisiting Knowledge Distillation via Label Smoothing Regularization}. In \bibinfo{booktitle}{\emph{CVPR}}. \bibinfo{pages}{3902--3910}.
\newblock


\bibitem[Zhang et~al\mbox{.}(2022)]%
        {zhang2022fine}
\bibfield{author}{\bibinfo{person}{Lin Zhang}, \bibinfo{person}{Li Shen}, \bibinfo{person}{Liang Ding}, \bibinfo{person}{Dacheng Tao}, {and} \bibinfo{person}{Ling{-}Yu Duan}.} \bibinfo{year}{2022}\natexlab{}.
\newblock \showarticletitle{Fine-tuning Global Model via Data-Free Knowledge Distillation for Non-IID Federated Learning}. In \bibinfo{booktitle}{\emph{CVPR}}. \bibinfo{pages}{10164--10173}.
\newblock


\bibitem[Zhu et~al\mbox{.}(2021)]%
        {zhu2021icml}
\bibfield{author}{\bibinfo{person}{Zhuangdi Zhu}, \bibinfo{person}{Junyuan Hong}, {and} \bibinfo{person}{Jiayu Zhou}.} \bibinfo{year}{2021}\natexlab{}.
\newblock \showarticletitle{Data-Free Knowledge Distillation for Heterogeneous Federated Learning}. In \bibinfo{booktitle}{\emph{ICML}}. \bibinfo{pages}{12878--12889}.
\newblock


\bibitem[Zhuang et~al\mbox{.}(2021)]%
        {DBLP:conf/mm/Zhuang0Z21}
\bibfield{author}{\bibinfo{person}{Weiming Zhuang}, \bibinfo{person}{Yonggang Wen}, {and} \bibinfo{person}{Shuai Zhang}.} \bibinfo{year}{2021}\natexlab{}.
\newblock \showarticletitle{Joint Optimization in Edge-Cloud Continuum for Federated Unsupervised Person Re-identification}. In \bibinfo{booktitle}{\emph{ACM Multimedia}}. \bibinfo{pages}{433--441}.
\newblock


\bibitem[Zhuang et~al\mbox{.}(2020)]%
        {10.1145/3394171.3413814}
\bibfield{author}{\bibinfo{person}{Weiming Zhuang}, \bibinfo{person}{Yonggang Wen}, \bibinfo{person}{Xuesen Zhang}, \bibinfo{person}{Xin Gan}, \bibinfo{person}{Daiying Yin}, \bibinfo{person}{Dongzhan Zhou}, \bibinfo{person}{Shuai Zhang}, {and} \bibinfo{person}{Shuai Yi}.} \bibinfo{year}{2020}\natexlab{}.
\newblock \showarticletitle{Performance Optimization of Federated Person Re-Identification via Benchmark Analysis}. In \bibinfo{booktitle}{\emph{ACM Multimedia}}. \bibinfo{pages}{955–963}.
\newblock
\showISBNx{9781450379885}


\end{thebibliography}

\end{document}